\documentclass[letterpaper]{article} %
\usepackage{aaai24}  %
\usepackage{times}  %
\usepackage{helvet}  %
\usepackage{courier}  %
\usepackage[hyphens]{url}  %
\usepackage{graphicx} %
\usepackage{natbib}  %
\usepackage{caption} %
\usepackage{algorithm}

\usepackage{newfloat}
\usepackage{listings}
\DeclareCaptionStyle{ruled}{labelfont=normalfont,labelsep=colon,strut=off} %
\floatstyle{ruled}
\newfloat{listing}{tb}{lst}{}
\floatname{listing}{Listing}
\usepackage{hyperref}
\usepackage{algpseudocode}
\usepackage{subfig}
\usepackage{enumitem}
\usepackage{tikz}
\usepackage{pifont}
\usepackage{amsmath}
\usepackage{amsthm}
\usepackage{amssymb}
\usepackage{booktabs}
\usepackage{float}
\usepackage{bm}
\usepackage{bbm}
\usepackage{multirow}
\usepackage{booktabs}
\usepackage{wrapfig}
\usepackage{amsthm,amsmath,amssymb}
\usepackage{mathrsfs}
\usepackage{algorithm}
\usepackage{color,xcolor}
\usepackage[english]{babel}
\newtheorem{example}{Example}
\newtheorem{theorem}{Theorem}

\newtheorem{lemma}{Lemma}
\newtheorem{objective}{Objective}
\usepackage[font=small]{caption}

\newcommand{\lbi}{\mathbf{x}^{\mathtt{lb}}_{i}}
\newcommand{\ulbi}{\mathbf{x}^{\mathtt{ulb}}_{i}}

\newcommand{\dulb}{\mathcal{D}^{\mathtt{ulb}}}
\newcommand{\gi}{g_{i,(c)}}
\newcommand{\ppi}{p_{i,(c)}}
\newcommand{\pwi}{\mathbf{p}^{\mathtt{w}}_{i}}
\newcommand{\lci}{|\mathcal{C}_{i}|}

\newcommand{\eg}{{\em e.g.}}           %
\newcommand{\ie}{{\em i.e.}}           %

\newcommand{\resb}[3]{#1$\pm$#2~(\textcolor{blue}{$\uparrow$#3})}

\newcommand{\tol}[3]{#1\ensuremath{_{\pm\mathtt{#2}}^{\textcolor{blue}{\uparrow\mathtt{#3}}}}}

\newcommand{\myboxc}[1]{\tikz[baseline=(MeNode.base)]{\node[rounded corners, fill=green!15](MeNode){#1};}}
\newcommand{\myboxb}[1]{\tikz[baseline=(MeNode.base)]{\node[rounded corners, fill=yellow!25](MeNode){#1};}}
\definecolor{light-gray}{gray}{0.6}
\definecolor{light-gray-2}{gray}{0.6}

\title{Roll With the Punches: Expansion and Shrinkage of \\Soft Label Selection for Semi-supervised Fine-Grained Learning}

\author {
    Yue Duan\textsuperscript{\rm 1}, 
    Zhen Zhao\textsuperscript{\rm 2}, 
    Lei Qi\textsuperscript{\rm 3}, 
    Luping Zhou\textsuperscript{\rm 2}, 
    Lei Wang\textsuperscript{\rm 4}, 
    Yinghuan Shi\textsuperscript{\rm 1}\thanks{Corresponding author.}
}
\affiliations {
    \textsuperscript{\rm 1}National Key Laboratory for Novel Software Technology, Nanjing University, China\\
    \textsuperscript{\rm 2}School of Electrical and Information Engineering, The University of Sydney, Sydney\\
    \textsuperscript{\rm 3}School of Computer Science and Engineering, Southeast University, China\\
    \textsuperscript{\rm 4}School of Computing and
    Information Technology, University of Wollongong, Australia\\
    yueduan@smail.nju.edu.cn, \{zhen.zhao, luping.zhou\}@sydney.edu.au, qilei@seu.edu.cn, leiw@uow.edu.au, syh@nju.edu.cn
}

\usepackage{bibentry}

\begin{document}

\maketitle

\begin{abstract}
While semi-supervised learning (SSL) has yielded promising results, the more realistic SSL scenario remains to be explored, in which the unlabeled data exhibits extremely high recognition difficulty, \eg, fine-grained visual classification in the context of SSL (SS-FGVC). The increased recognition difficulty on fine-grained unlabeled data spells disaster for pseudo-labeling accuracy, resulting in poor performance of the SSL model. To tackle this challenge, we propose Soft Label Selection with Confidence-Aware Clustering based on Class Transition Tracking (SoC) by reconstructing the pseudo-label selection process by jointly optimizing Expansion Objective and Shrinkage Objective, which is based on a soft label manner. Respectively, the former objective encourages soft labels to absorb more candidate classes to ensure the attendance of ground-truth class, while the latter  encourages soft labels to reject more noisy classes, which is theoretically proved to be equivalent to entropy minimization. In comparisons with various state-of-the-art methods, our approach demonstrates its superior performance in SS-FGVC. 
   Checkpoints and source code are available at \url{https://github.com/NJUyued/SoC4SS-FGVC}.
\end{abstract}

\section{Introduction}
\label{sec:intro}

Semi-supervised learning (SSL) aims to leverage a pool of unlabeled data to alleviate the dependence of deep models on labeled data \cite{zhu2009introduction}. However, current SSL approaches achieve promising performance with clean and ordinary data, but become unstuck against the indiscernible
 unlabeled data.
A typical and worth discussing example is the semi-supervised fine-grained visual classification (SS-FGVC) \cite{su2021realistic}, where the unlabeled data faced by the SSL model is no longer like the ``house'' and ``bird'' that are easy to distinguish, but like the dizzying ``Streptopelia chinensi'' and ``Streptopelia orientalis'', which are difficult to distinguish accurately even for ornithologists.
The mentioned scenario also hints at the practicality of SS-FGVC, \ie, it is resource-consuming to label the fine-grained data for supervised learning.

\begin{figure}[t]
  \centering
  \subfloat[Semi-Aves]{
  \includegraphics[width=4cm,height=3.5cm]{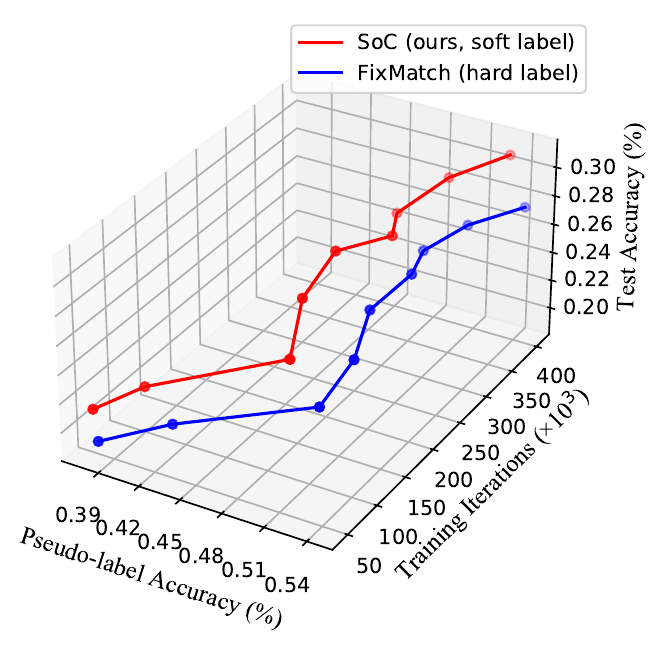}
  \label{fig:tab34-a}
  }
  \hspace{-3mm}
  \subfloat[Semi-Fungi]{
  \includegraphics[width=4cm,height=3.5cm]{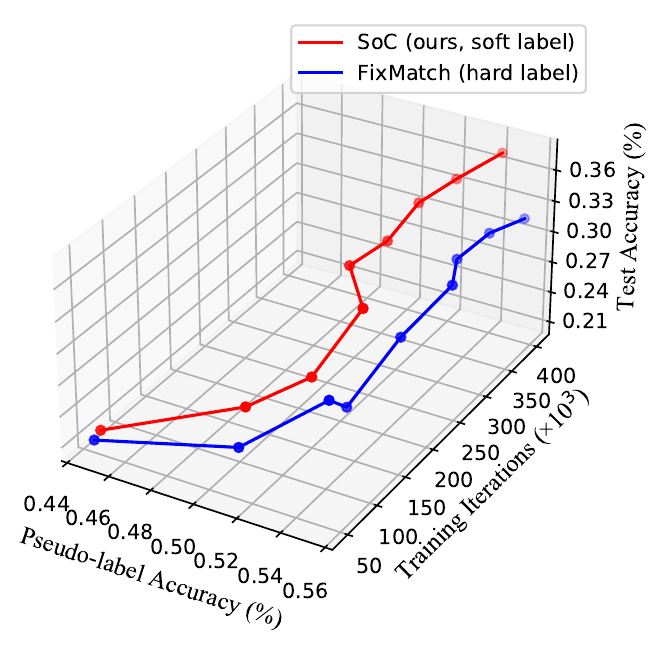}
  \label{fig:tab34-c}
  }
  \caption{Experimental results on Semi-Aves and Semi-Fungi (see Sec. \ref{sec:dataset}). Our SoC (soft label) consistently outperforms FixMatch (hard label) even when using pseudo-labels with lower accuracy.}
  \label{fig:intro2}
  \vskip -0em
\end{figure}

\begin{figure}[t] 
    \centering
    \includegraphics[width=0.9\linewidth]{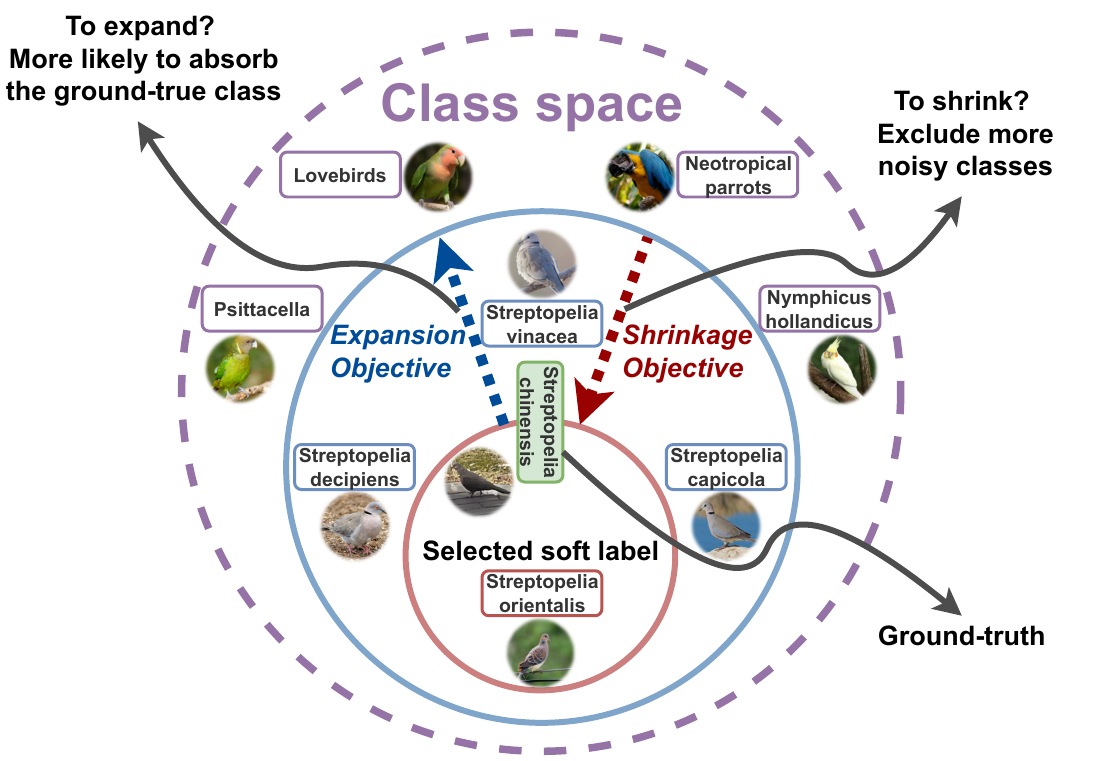}

    \caption{Illustration for proposed approach of soft label selection. We divide the class space of the SS-FGVC scenario into clusters with different granularity, where each cluster (\eg, the \textcolor[RGB]{0,76,153}{blue circle} and \textcolor[RGB]{153,0,0}{red circle}) contains classes that are more similar to each other. We encourage the soft pseudo-label to select the classes in a cluster with smaller granularity and higher probability of containing ground-truth (\ie, \textcolor
   [RGB]{130,179,102}{``Streptopelia chinensis''}), by optimizing  \textcolor[RGB]{0,76,153}{\textit{Expansion Objective}} (to absorb more candidate classes) and \textcolor[RGB]{153,0,0}{\textit{Shrinkage Objective}} (to shrink the cluster for rejecting noisy classes). }
    \label{fig:intro}
  \end{figure}

SS-FGVC scenario poses a more problematic problem for pseudo-labeling-based SSL approaches. In general, pseudo-labeling \cite{lee2013pseudo,iscen2019label,arazo2020pseudo} based methods assign pseudo-labels to the unlabeled data online for self-training loops, where these methods rely heavily on the accuracy of pseudo-labels \cite{xie2020unsupervised,sohn2020fixmatch,li2021comatch,zhang2021flexmatch,zhao2022dc}. Unfortunately, fine-grained data may severely affect the quality of pseudo-labels and consequently  pull down the model performance. As an example, FixMatch \cite{sohn2020fixmatch}, currently the most popular SSL method, selects high-confidence pseudo-labels by a threshold, which are more likely correct in intuition. Then the selected pseudo-labels are converted into one-hot labels for training. If applying to SS-FGVC, the low accuracy of pseudo-labels caused by fine-grained data will pose a great risk to the model training, resulting in further error accumulation in subsequent iterations. Therefore, considering the difficulty of accurately predicting the pseudo-labels in the SS-FGVC scenario, why not step back on a relaxed manner to generate pseudo-labels? A natural idea is to use the soft label. 
Our motivation is that training with proper soft labels is more robust to 
noisy pseudo-labels compared to hard labels in SS-FGVC. As shown in Fig. \ref{fig:intro2}, The proposed SoC (as described below) unexpectedly maintains a performance advantage during training, even when using soft labels with lower pseudo-label accuracy. The bad effects of incorrect hard label on the model have overridden its low entropy advantage while soft label can benefit the model because it could still provide useful information although it is wrong. 
However, using the traditional soft label may robbing Peter to pay Paul, because it contains all the classes, which obviously introduces too much noise into the learning. 
We therefore revisit the soft label with the purpose of reducing the penalty to wrong pseudo-labels while preventing the noisy classes in the label from confusing the model. 
Based on this, we target to find a subset of the class space for each soft pseudo-label, which contains the ground-truth classes as much as possible and the noisy classes as few as possible. Thus, as illustrated in Fig. \ref{fig:intro}, we propose the soft label selection with a coupled optimization goal to include the classes that are more likely to be ground-truth, and to exclude the classes that are more likely to be noise. Specifically, for \textbf{\textit{Expansion Objective}} (Obj. \ref{eq:ob1}), the classes that are most similar to the current class prediction should be treated as the candidates for the ground-truth; For \textbf{\textit{Shrinkage Objective}} (Obj. \ref{eq:ob2}), the more confident the model is in its prediction, the more it should shrink the range of candidate classes.

\newcommand{\mzz}{3.9cm}
\newcommand{\widd}{4cm}

\begin{figure*}[t]  
   \begin{center}
   \resizebox{0.9\linewidth}{!}{          
      \includegraphics{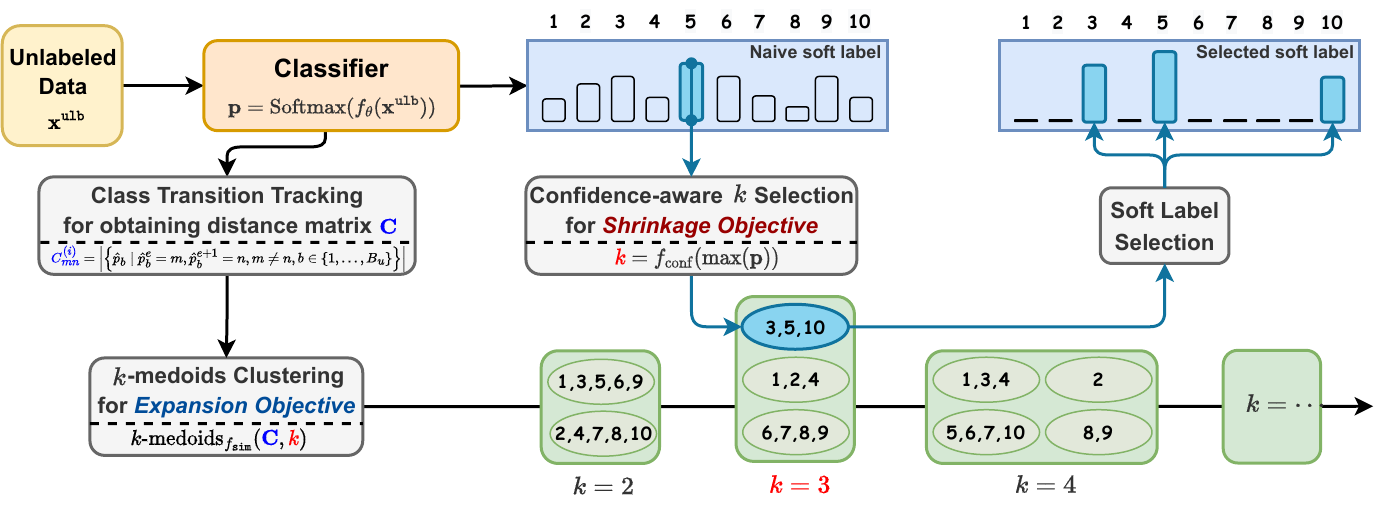}
   }  
   \end{center}
   \caption{Overview of SoC. We select a subset of the class space to serve as the selection range of candidate classes, encouraging the attendance of ground-truth class as much as possible (Obj. \ref{eq:ob1}) while rejecting noisy class as much as possible (Obj. \ref{eq:ob2}). Given the unlabeled samples, we perform Class Transition Tracking (Eq. \eqref{eq:track}) to count the transitions of class predictions to obtain the similarity between classes (see Sec. \ref{sec:cttc}). With obtained similarity, we perform $k$-medoids clustering on the class space to obtain the clusters of candidate classes, which is used to select the soft pseudo-labels (Eq. \eqref{eq:km} Eq. \eqref{eq:s}). 
   For different samples, we decide the selection of $k$ based on the confidence scores of their class predictions (Eq. \eqref{eq:conf}), where  higher confidence corresponds to a larger $k$, \ie, a smaller selection range of candidate classes. }

   \label{fig:soc}
\end{figure*}

Given aforementioned discussions, we propose \textbf{So}ft Label Selection with Confidence-Aware \textbf{C}lustering based on Class Transition Tracking (\textbf{SoC}) to establish a high-performance SS-FGVC framework, which can be decomposed into two parts: (1) To optimize \textbf{\textit{Expansion Objective}}: Overall, we utilize a $k$-medoids-like clustering algorithm \cite{kaufman2009finding} to serve as a class selector for the soft pseudo-label generation. In order to make soft labels more likely to contain the ground-truth class, when the model gives a class prediction for an unlabeled sample, other classes similar to it, \ie, all classes in a cluster, are taken into account. The key of clustering on the class space is the distance metric between classes. However, Euclidean distances usually used in classic $k$-medoids algorithm will have difficulty in directly representing class-to-class distances.
Thus, we innovatively introduce Class Transition Tracking (CTT) technique to measure the similarity (rather than distance) between classes for clustering ($k$-medoids algorithm can directly accept similarity as input). For specific, we statistically track the transitions of the model's class predictions for each sample, \eg, for the same unlabeled data, the model predicts ``Streptopelia chinensis'' for it in one epoch and ``Streptopelia
orien'' for it in the next epoch. This oscillation of predictions indicates the degree of similarity between classes, \ie, the more frequent the class transition, the more similar the two classes are and the closer they are in the class space. With CTT, the $k$-medoids clustering can be performed to select the candidates of soft label for the unlabeled data. (2) To optimize \textbf{\textit{Shrinkage Objective}}: In a nutshell, we shrink the obtained clusters based on the confidence scores of predictions on the unlabeled data. We theoretically prove that in SoC, shrinking the soft labels could lead to entropy minimization \cite{grandvalet2005semi}. Thus, we use the number of clusters (called $k$) to control the granularity of clustering, where $k$ is mapped by the confidence, \ie,  certain predictions should correspond to a smaller extension range (\ie, selecting smaller $k$), and conversely, the soft label should be further extended for uncertain predictions (\ie, selecting larger $k$). The overview of SoC is shown in Fig. \ref{fig:soc}.

In summary, our contributions are as follows: (1) We propose a coupled optimization goal for SS-FGVC, which is composed of \textbf{\textit{Expansion Objective}} and \textbf{\textit{Shrinkage Objective}};
(2) We propose soft label selection framework with optimizing the above objectives by Class Transition Tracking based $k$-medoids clustering and Confidence-Aware $k$ Selection; 
(3) Finally, SoC shows promising performance in SS-FGVC under both the conventional setting and more realistic setting, outperforming various state-of-the-art methods.

\section{Related Work}
\label{sec:rework}
\textit{Semi-supervise learning} (SSL) relies on leveraging the unlabeled samples to boost the performance of model trained on the labeled data \cite{gong2016multi,van2020survey,li2022pln,xing2022semi,Xing_2023_CVPR,zhao2023instance,duan2023towards,yang2023shrinking}. For semi-supervised image classification, recently proposed SSL approaches unify \textit{pseudo-labeling} technique to assign pseudo-labels to unlabeled data, enabling their use in the training.
The pseudo-label can be classified into two classes: the ``hard'' label (one-hot label) used in \citet{lee2013pseudo,sohn2020fixmatch,zhang2021flexmatch} and the ``soft'' label used in \citet{xie2020unsupervised,berthelot2020remixmatch,li2021comatch,zheng2022simmatch}, while the training on pseudo-labels also can be classified into two classes: training multiple models to exploit the disagreement among them in \citet{qiao2018deep,dong2018imbalanced}; and training on the labeled data to obtain pseudo-labels on the unlabeled data for subsequent training \cite{lee2013pseudo,sohn2020fixmatch,li2021comatch,zhang2021flexmatch,duan2022mutexmatch}. 

In the conventional SSL, the datasets is coarse-grained (\eg,  ``automobile'', ``cat'' in CIFAR-10 \cite{krizhevsky2009learning}) while datasets in the real world usually contain fine-grained data. Although  \textit{fine-grained visual classification} (FGVC) is widely discussed \cite{biederman1999subordinate,dubey2018pairwise,zhang2018fine,wei2021fine,nassar2021all}, FGVC in the context of SSL (SS-FGVC) is always ignored. Recently, \citet{su2021realistic} takes the lead in evaluating the SS-FGVC setting. 
In this work, we propose that the pseudo-labeling scheme could be performed in a relaxed manner 
with a coupled optimization goal, which reaches superior performance cross a variety of SS-FGVC scenarios.

\section{Methodology}
\subsection{A Coupled Optimization Goal for SS-FGVC}
\label{sec:md}
SSL involves a training dataset $\mathcal{D}$ divided into two portions: the set of labeled data $\mathcal{D}^{\mathtt{lb}} =\{(\mathbf{x}_{i}^{\mathtt{lb}},\mathbf{y}_{i})\}^{N}_{i=1}$ and the set of unlabeled data $\mathcal{D}^{\mathtt{ulb}}=\{\mathbf{x}_{i}^{\mathtt{ulb}}\}^{M}_{i=1}$, where $N$ and $M$ are the amount of samples from $K$ classes in $\mathcal{D}^{\mathtt{lb}}$ and $\mathcal{D}^{\mathtt{ulb}}$ respectively. Formally, $\mathbf{x}$ is sampled from distribution $\mathcal{X}$ while $\mathbf{y}\in\mathcal{Y}=\{1,\dots,K\}$. 
SSL algorithms aim to learn a predictor $f_{\theta}: \mathcal{X}\rightarrow\mathcal{Y}$ to correctly classify the test data, where $\theta$ is the model parameters.
In SS-FGVC, $\mathcal{X}$ presents fine-grained similarity between classes, which poses new challenges of identifying sub-classes that belong to a general class. 

Current conventional SSL methods incorporate pseudo-labeling to achieve entropy minimization for the predictions on unlabeled data $\ulbi$. Precisely, given $\ulbi$ in the self-training loop, the model computes its output probability $\mathbf{p}_{i}=\mathrm{Softmax}(f_{\theta}(\ulbi))$ and utilizes the hard pseudo-label $\hat{p}_{i}=\arg\max(\mathbf{p}_{i})$ to augment the training set by ${\mathcal{D}}'=(\mathcal{D}\setminus\{\ulbi\})\cup\{(\ulbi,\hat{p}_{i})\}$. In this way, the model is able to leverage both labeled and unlabeled data to produce a robust learning scheme to improve the performance of SSL. Mathematically, the common goal of pseudo-labeling based methods is to minimize the following objective:

\begin{equation}
\resizebox{\hsize}{!}{
$\begin{aligned} \mathcal{L} =\frac{1}{N}\sum_{i=1}^{N} H(f_{\theta}(\mathbf{x}^{\mathtt{lb}}_{i}),\mathbf{y}^{\mathtt{lb}}_{i})+
\lambda\mathbbm{1}(\max(\mathbf{p}_{i})\geq\tau)\frac{1}{M}\sum_{i=1}^{M} H(f_{\theta}(\mathbf{x}^{\mathtt{ulb}}_{i}),\hat{p}_{i})
\end{aligned}$
}
\label{eq:plm}
\end{equation}

where $H(\cdot,\cdot)$ is the standard cross entropy, $\lambda$ is the trade-off coefficient, $\mathbbm{1}(\cdot)$ is the indicator function and $\tau$ is a pre-defined threshold to filter out low-confidence samples. Pseudo-labeling achieves remarkable performance in conventional SSL by reducing the entropy of predictions, \ie, 
\begin{equation}
    \mathop{\min}_{\theta} \mathbbm{E}_{\ulbi\in\dulb}[\mathcal{H}(\mathbf{p}_{i})],\label{eq:em}
\end{equation}where $\mathcal{H}(\cdot)$ refers to the entropy. However, there are potential risks in using this technique in SS-FGVC. Given that the difficulty of the model to correctly identify fine-grained images is not comparable to the difficulty of correctly identifying regular images, the resulting hard labels are mixed with too much noise. The damage caused to the model by a wrong hard label is nonnegligible, which means that pseudo-labeling could not work well when the model can generate pseudo-labels with low accuracy. Unfortunately, it is difficult to distinguish fine-grained data easily in SS-FGVC. Therefore, it is natural to consider the use of soft labels in SS-FGVC to participate in the self-training process. Meanwhile, we still tend to achieve entropy minimization to some extent, which is involved in Obj. \ref{eq:ob2}.

The core idea of SoC is that the selected soft label for each $\ulbi$ correspondingly contains only a roper subset $\mathcal{C}_{i}$ of $\mathcal{Y}$, aiming at reducing the noise present in pseudo-label to yield a superior performance. Formally, we denote the $i$-th component of vector $v$ as $v_{(i)}$. Hereafter, \textit{label selection indicator} is defined as $\mathbf{g}_{i}\subset\{0,1\}^{K}$, which is a binary vector representing whether a class is selected or not for the pseudo-label $\mathbf{p}$ on $\ulbi$, \ie, $g_{i,(c)}=1$ indicates $p_{i,(c)}$ is selected while $g_{i,(c)}=0$ indicates $p_{i,(c)}$ is not selected. Specifically, we compute $g_{i,(c)}$ by
\begin{equation}
    g_{i,(c)}=\mathbbm{1}(c\in \mathcal{C}_{i}), 
    \label{eq:gi}
\end{equation}
where we will describe later how $\mathcal{C}_{i}$ is obtained. With obtained $\mathbf{g}_{i}$, we select pseudo-label $\mathbf{p}_{i}$ by
\begin{equation}
    \Tilde{\mathbf{p}}_{i}=\mathrm{Normalize}(\mathbf{g}_{i}\circ\mathbf{p}_{i}),\label{eq:p}
\end{equation}
where $\mathrm{Normalize}(x)_{i}=x_{i}/{\sum_{j=1}^{K}x_{j}}$ is normalization operation and $\circ$ is Hadamard product. Then $\Tilde{\mathbf{p}}_{i}$ is used for training with optimizing our \textit{coupled goal for SS-FGVC}.

\begin{objective}[Expansion Objective]
\label{eq:ob1}
Encourage the pseudo-label to contain the ground-truth class as much as possible:
\begin{equation}
    \mathop{\max}_{\theta} \mathbbm{E}_{\ulbi\in\dulb}\left[ \mathbbm{1} (y^{\star}_{i}\in \left\{c\mid \gi=1\right\})p_{i,(y^{\star}_{i})} \right], 
\end{equation}
where $y^{\star}_{i}\in \mathcal{Y}$ is the ground-truth label of $\ulbi$. For simplicity, we denote $p_{i,(y^{\star}_{i})} \mathbbm{1} (y^{\star}_{i}\in \left\{c\mid \gi=1\right\})$ as $z^{\mathtt{obj1}}_{i}$.
\end{objective}
Intuitively, given that the soft label generated by SoC does not contain all classes (\ie, $\mathcal{C}_{i}\subsetneqq\mathcal{Y}$), a necessary prerequisite for $\Tilde{\mathbf{p}}_{i}$ to be useful for training is that $y^{\star}_{i}$ is selected into the pseudo-label. This objective corresponds to our more relaxed soft-label training scheme compared to the pseudo-labeling approaches. An extreme way to optimize this objective is to incorporate as many classes as possible into the pseudo-label, but this introduces more noisy information. Another extreme way to optimize this objective is to only select the semantic class of $\ulbi$ (\ie, the correct hard pseudo-label), but it is very difficult for SS-FGVC to reach this point. Thus, we should balance the advantages and disadvantages to optimize the first objective, 
\ie, we suppress the tendency to make soft labels contain all classes.

\begin{objective}[Shrinkage Objective]
\label{eq:ob2}
Encourage the pseudo-label to contain as few classes as possible:
\begin{equation}
    \mathop{\min}_{\theta} \mathbbm{E}_{\ulbi\in\dulb}\sum_{c=1}^{K} \gi, 
\end{equation}
where $\sum_{c=1}^{K} \gi$ is denoted as $z^{\mathtt{obj2}}_{i}$ for simplicity.
\end{objective}
Briefly, we optimize the first goal while minimizing the number of classes in soft labels, \ie, shrinking the range where the ground-truth class may exist to exclude noisy classes. In SoC, this objective corresponds to a mathematical interpretation: the familiar entropy minimization objective \cite{grandvalet2005semi}, \ie, Eq. \eqref{eq:em} (detailed proof is shown in Sec. \ref{sec:cas}). 

Given $y^{\star}_{i}$ is unseen in SSL, we first introduce how to optimize Obj. \ref{eq:ob1} heuristically in the following subsection.

\subsection{Class Transition Tracking based Clustering}
\label{sec:cttc}
Notably, the key to optimizing Obj. \ref{eq:ob1} is to obtain a suitable $\mathcal{C}_{i}$ that contains the semantic class of the unlabeled data as much as possible. The starting point of our solution is: \textit{similar classes are more likely to be misclassified, and the correct answer is hidden among them with a high probability}. Thus, given $\ulbi$, when the model considers it to be a ``streptopelia chinensis'', those classes that are most similar to ``streptopelia chinensis'' are also likely to be the ground-truth class (\eg, ``streptopelia orientalis''). A reliable solution is to cluster around ``streptopelia chinensis'', and then we add the classes in the cluster to $\mathcal{C}_{i}$. For simplicity, we adopt $k$-medoids clustering algorithm \cite{kaufman2009finding} in SoC to obtain $\mathcal{C}_{i}$. 
It is well known that $k$-medoids clustering can accept distance or similarity as input.
Because we aim to cluster the classes instead of the samples, it is not convenient to use Euclidean distance, which is most commonly used in $k$-medoids clustering, to achieve our goal.
Hence, we innovatively propose class transition tracking (\textbf{CTT}) to simply and effectively model the similarity between fine-grained classes. 

In the training, the class prediction of the SSL model for a given sample $\ulbi$ is not constant. As new knowledge is learned, the class predictions output by the model may transit from one class to another, \ie, $\hat{p}_{i}^{e}=m$ at epoch $e$ transits to $\hat{p}_{i}^{e+1}=n$ at epoch $e+1$, where $m$ and $n$ are different classes. First, a adjacency matrix  $\mathbf{C}\in\mathbb{R}^{K\times K}_{+} $ is constructed, where each element $C_{mn}$ represents the frequency of class transitions that occur from class $m$ to class $n$. $C_{mn}$ is parametrized by the following CTT averaged on last $N_{B}$ batches with unlabeled data batch size $B_{u}$, \ie, $C_{mn}=\sum_{i=1}^{N_{b}}{C^{(i)}_{mn}}/N_{b}$, where 
\begin{equation}
\resizebox{\hsize}{!}{
$\begin{aligned} C^{(i)}_{mn}= 
  \left| \left\{\hat{p}_{b}\mid\hat{p}^{e}_{b}=m,\hat{p}^{e+1}_{b}=n,m\neq n,b\in  \left\{1,...,B_{u}\right\}\right\} \right|. 
\end{aligned}$
}
 \label{eq:track}
\end{equation}

Our core idea is: \textit{the more frequent the transition between two classes, the greater the similarity between the two classes and the closer they are} (see Sec. \ref{app:eec} for more empirical evaluations on this CTT-based similarly measure properties).
Thus, we treat $f_{\mathtt{sim}}(m,n)=\frac{C_{mn}+C_{nm}}{2}$ as a similarity function, \ie, the larger the $f_{\mathtt{sim}}(m,n)$, the more similar the two classes $m$ and $n$ are. This measurement of image semantic similarity based on the model itself is more discriminative. Finally, we plug $f_{\mathtt{sim}}(m,n)$ into a $k$-medoids-like clustering process to obtain the class clusters:
\begin{equation}
    \mathcal{S}=k\text{-}\mathrm{medoids}_{f_{\mathtt{sim}}}(\mathbf{C},k),
    \label{eq:km}
\end{equation}

where $\mathcal{S}=\left\{ S_{c} \mid c\in\{1,\dots,k\} \land S_{c}\subsetneqq\mathcal{Y} \right\}$ is obtained $k$ clusters. Algorithmic presentation and computational cost analysis can be respectively found in Algorithm \ref{a} and Sec. \ref{app:cc} of supplementary. Then, we construct $\mathcal{C}_{i}$ by
\begin{equation}
    \mathcal{C}_{i}=S_{s}, s=\mathop{\arg\max}_{c\in\{1,\dots,k\}}\mathbbm{1}( \hat{p}_{i} \in S_{c})
    \label{eq:s}
\end{equation}
which means we select the cluster containing the most confident class prediction outputted by the model as $C_{i}$. In this way, although we cannot explicitly optimize Obj. \ref{eq:ob1} (since we can never know the ground-truth labels of the unlabeled data), we can optimize it heuristically, since the clusters generated based on CTT contain families of classes that are most likely to be misclassified by the classifier, and in which are more likely to contain the ground-truth class.

\subsection{Confidence-Aware $k$ Selection}
\label{sec:cas}
As mentioned in Sec. \ref{sec:md}, we first show that optimizing Obj. \ref{eq:ob2} is equivalent to entropy minimization (\ie, Eq. \eqref{eq:em}), which is a widely used objective for effectively improving the quality of pseudo-labels \cite{miyato2019virtual,sohn2020fixmatch}. Denoting $\mathcal{C}_{i}$ at epoch $e$ as $\mathcal{C}^{(e)}_{i}$, the following holds:
\begin{theorem}
\label{the2}
In SoC, minimizing $|\mathcal{C}^{(1)}_{i}|$ to $|\mathcal{C}^{(m)}_{i}|$:  $\mathcal{C}^{(m)}_{i}\subsetneqq\dots\mathcal{C}^{(2)}_{i}\subsetneqq\mathcal{C}^{(1)}_{i}$ (\ie, $\sum_{c=1}^{K} \mathbf{g}^{(m)}_{i}<\dots<\sum_{c=1}^{K}\mathbf{g}^{(2)}_{i}<\sum_{c=1}^{K} \mathbf{g}^{(1)}_{i}$), we show that the entropy of $\mathbf{p}_{i}$ is minimizing:
\begin{equation}
    \mathcal{H}(\Tilde{\mathbf{p}}^{(m)}_{i})\leq\mathcal{H}(\Tilde{\mathbf{p}}^{(m-1)}_{i})\leq\dots \mathcal{H}(\Tilde{\mathbf{p}}^{(2)}_{i})\leq \mathcal{H}(\Tilde{\mathbf{p}}^{(1)}_{i}),
    \label{eq:th2}
\end{equation}
where $\mathcal{H}(\cdot)$ refers to the entropy.
\end{theorem}
To prove this theorem, we first give the following lemma.
\begin{lemma}
\label{the}
Given $\Tilde{\mathbf{p}}_{i}$ in SoC  (implying $\mathcal{C}_{i}\subsetneqq\mathcal{Y}$ and $\hat{p}_{i}\in\mathcal{C}_{i}$), we show that the entropy of $\Tilde{\mathbf{p}}_{i}$ is smaller than that of $\mathbf{p}_{i}$:
\begin{equation}
    \mathcal{H}(\Tilde{\mathbf{p}}_{i})\leq \mathcal{H}(\mathbf{p}_{i}),
    \label{eq:the}
\end{equation}
where $\mathcal{H}(\cdot)$ refers to the entropy.
\end{lemma}

See Sec. \ref{sec:proof} of supplementary for proof. 
By Lemma \ref{the}, we can simply obtain the proof for Theorem \ref{the2}.
\begin{proof}[Proof for Theorem 1.] 
Given $\mathcal{C}^{(2)}_{i}\subsetneqq\mathcal{C}^{(1)}_{i}$, we can treat $\Tilde{\mathbf{p}}^{(1)}_{i}$ as $\mathbf{p}_{i}$ and $\Tilde{\mathbf{p}}^{(2)}_{i}$ as $\Tilde{\mathbf{p}}_{i}$ in Lemma \ref{the}, \ie, treat the previously obtained $\Tilde{\mathbf{p}}_{i}$ as the naive soft pseudo-label. 
Thus, $\mathcal{H}(\Tilde{\mathbf{p}}^{(2)}_{i})\leq\mathcal{H}(\Tilde{\mathbf{p}}_{i}^{(1)})$ holds. 
For any $\mathcal{C}^{(j)}_{i}\subsetneqq\mathcal{C}^{(j-1)}_{i}$, we can repeat the above proof to obtain $\mathcal{H}(\Tilde{\mathbf{p}}_{i}^{(j)})\leq\mathcal{H}(\Tilde{\mathbf{p}}_{i}^{(j-1)})$. Thus, we can obtain $\mathcal{H}(\Tilde{\mathbf{p}}^{(m)}_{i})\leq\dots\leq\mathcal{H}(\Tilde{\mathbf{p}}^{(1)}_{i})$.
\end{proof}

By Theorem \ref{the2}, 
we propose to ``shrink'' $\mathcal{C}_{i}$ to optimize Obj. \ref{eq:ob2}. Since $\mathcal{C}_{i}$ is obtained by Eq. \eqref{eq:s}, a feasible way to ``shrink'' $\mathcal{C}_{i}$ is to decrease the number of classes in the clusters obtained by $k$-medoids clustering, \ie, enlarging $k$. An extreme case is setting $k=K$, which means each cluster only contains one class, \ie, $|\mathcal{C}_{i}|=1$. Meanwhile, with $\mathcal{C}_{i}$ obtained in this way, there is only one component in $\mathbf{g}_{i}$ that is not zero, which means the pseudo-label $\mathbf{p}_{i}$ generated by SoC will also degrade to one-hot labels (by Eq. \eqref{eq:p}). With training, the model will become more confident in $\ulbi$, and therefore we should shrink the range of candidate classes (\ie, $|\mathcal{C}_{i}|$) for $\ulbi$ to optimize Obj. \ref{eq:ob2}, \ie, the larger the $\max(\mathbf{p}_{i})$, the finer the clustering granularity, which implies the larger $k$. Thus, we propose Confidence-Aware $k$ Selection: given $\ulbi$, $k_{i}$ is calculated as
\begin{equation}
    k_{i}=f_{\mathrm{conf}}(\max(\mathbf{p}_{i})),\label{eq:conf}
\end{equation}
where $f_{\mathrm{conf}}(\cdot)$ is a monotonically increasing function that maps confidence scores to $k$. In SoC, the linear function is adopted as $f_{\mathrm{conf}}$ for simplicity:
\begin{equation}
    k_{i}=\lceil(\frac{\max(\mathbf{p}_{i})}{\alpha}+\frac{2}{K}) \times K -\frac{1}{2} \rceil,
    \label{eq:k}
\end{equation}
where $\alpha\geq\frac{K}{K-2}$ is a pre-defined parameter. By this, we control $k_{i}\in\{2,\dots,K\}$. More discussion on other adopted functions for $f_{\mathrm{conf}}$ can be found in Sec. \ref{sec:ab}.

Additionally, the dynamic selection of $k$ corresponds to another intuitive motivation. It is worth noting that since the identification difficulty of different samples is different, it is not reasonable that all samples' $\mathcal{C}$ are obtained from clusters of the same granularity (\ie, $k$ is fixed). Taking the above considerations into account, we believe that the clustering granularity should be determined at the sample level. The simpler samples should be from clusters with finer granularity (\ie, larger $k$). Accordingly, the harder samples should be from clusters with coarser granularity (\ie smaller $k$), which implies optimizing Obj. \ref{eq:ob1} since a larger $|\mathcal{C}_{i}|$ will have a larger probability to contain the ground-truth class of $\ulbi$.
For simplicity, we regard the prediction's confidence score $\max(\mathbf{p}_{i})$ as an estimation of the learning difficulty of $\ulbi$. This also corresponds to our $k$ selection scheme: the larger the confidence, the larger the $k$.

\subsection{Putting It All Together}
Following prevailing SSL methods \cite{xie2020unsupervised,sohn2020fixmatch,li2021comatch,zhang2021flexmatch,tai2021sinkhorn}, \textit{consistency regularization} technique is integrated into SoC, \ie, weak augmentation $\mathrm{Aug}_{\mathtt{w}}(\cdot)$ is applied on $\lbi$ and $\ulbi$ while strong augmentation $\mathrm{Aug}_{\mathtt{s}}(\cdot)$ is only applied on $\ulbi$. In a training iteration, we obtain a batch of $B$ labeled data $\{(\lbi,\mathbf{y}_{i})\}^{B}_{i=1}$ and a batch of $\mu B$  unlabeled data $\{\ulbi\}^{\mu B}_{i=1}$, where $\mu$ determines the relative size of labeled batch to unlabeled batch. First, the supervised loss $\mathcal{L}_{\mathtt{sup}}$ is defined as
\begin{equation}
\mathcal{L}_{\mathtt{sup}}=\frac{1}{B}\sum_{n = 1}^{B}H(\mathbf{y}_{i},f_{\theta}(\mathrm{Aug}_{\mathtt{w}}(\ulbi))),
\label{eq:sup} 
\end{equation}
where $H(\cdot,\cdot)$ denotes the cross-entropy loss. Then, the whole algorithm proceeds by three steps:

\noindent\textbf{Step 1.} The soft pseudo-label for weakly-augmented $\ulbi$ is computed as $\pwi=\mathrm{Softmax}(f_{\theta}(\mathrm{Aug}_{\mathtt{w}}(\ulbi)))$. With obtained $\pwi$, the $k_{i}^{\mathtt{w}}$ for clustering is selected by Eq. \eqref{eq:k}. 

\noindent\textbf{Step 2.} Class transition matrix $\mathbf{C}$ is calculated by Eq. \eqref{eq:track} in the current iteration. By Eq. \eqref{eq:km}, the CTT based $k$-medoids clustering is performed with $k_{i}^{\mathtt{w}}$ and $\mathbf{C}$ to obtain $\mathcal{C}_{i}^{\mathtt{w}}$.

\noindent\textbf{Step 3.} $\mathbf{g}_{i}^{\mathtt{w}}$ is computed by Eq. \eqref{eq:gi} with $\mathcal{C}_{i}^{\mathtt{w}}$ and then the selected pseudo-label $\Tilde{\mathbf{p}}_{i}^{\mathtt{w}}$ is computed by Eq. \eqref{eq:p} with $\mathbf{g}_{i}^{\mathtt{w}}$.

Finally, the consistency regularization is achieved by minimizing the consistency loss $\mathcal{L}_{\mathtt{cos}}$:\begin{equation}
\mathcal{L}_{\mathtt{cos}}=\frac{1}{\mu B}\sum_{i = 1}^{\mu B}H(\Tilde{\mathbf{p}}_{i}^{\mathtt{w}},f_{\theta}(\mathrm{Aug}_{\mathtt{s}}(\ulbi))).
\label{eq:cr} 
\end{equation}
Unlike previous prevailing pseudo-labeling based methods (\eg, \citet{sohn2020fixmatch,li2021comatch}) that waste data with low confidence (as shown in Eq. \eqref{eq:plm}), SoC exploits all unlabeled data for training. The total loss function is given by:
\begin{equation}
\mathcal{L} =\mathcal{L}_{\mathtt{sup}}+\lambda_{\mathtt{cos}} \mathcal{L}_{\mathtt{cos}},
\end{equation}
where $\lambda_{\mathtt{cos}}$ is a hyper-parameter to weight the importance of the consistency loss. So far, we have established the framework of \textbf{So}ft Label Selection with CTT based $k$-medoids \textbf{C}lustering (\textbf{SoC}) which is boosted by Confidence-Aware $k$ Selection to address SS-FGVC. The whole algorithm is presented in Algorithm \ref{a2} of supplementary.
\begin{table*}[t]

   \caption{Accuracy (\%) on Semi-Aves and Semi-Fungi. We provide comparisons with multiple baseline methods reported in \citet{su2021realistic} and the state-of-the-art SSL methods based on our re-implementation (marked as $^\ast$). The models are trained from scratch, or ImageNet/iNat pre-trained or the model initialized with MoCo learning on the unlabeled data. Our results are averaged on 3 runs while the standard deviations $\pm\mathtt{Std.}$ are reported. We respectively report the percentage difference in performance between SoC and the \protect\myboxb{best SSL baselines}, as well as between SoC+MoCo and the \protect\myboxc{best MoCo baselines}. Meanwhile, we mark out the \textit{best SSL results} and the \textbf{best MoCo results}.}
   \vspace{-0.5em}
    \centering
  \resizebox{\linewidth}{!}{  
      \label{table:1}
      \begin{tabular}{@{}c|c|c|c|cc|cc|cc@{}}     
      \toprule\toprule
     \multirow{2}{*}{Dataset} & \multirow{2}{*}{Pseudo-label}  & \multirow{2}{*}{Method} & \multirow{2}{*}{Year} & \multicolumn{2}{c|}{from scratch} & \multicolumn{2}{c|}{from ImageNet} & \multicolumn{2}{c}{from iNat}  \\  
      
       & & & & Top1 &Top5          & Top1 &Top5 & Top1 &Top5   \\\midrule 
      \multirow{13}{*}{Semi-Aves} & \multirow{2}{*}{---}   &\textcolor{light-gray-2}{Supervised oracle} & --- &\textcolor{light-gray-2}{57.4$\pm$0.3 }&\textcolor{light-gray-2}{79.2$\pm$0.1 }&\textcolor{light-gray-2}{68.5$\pm$1.4 }&\textcolor{light-gray-2}{88.5$\pm$0.4 }&\textcolor{light-gray-2}{69.9$\pm$0.5 }&\textcolor{light-gray-2}{89.8$\pm$0.7}\\  
     & & MoCo \cite{he2020momentum}& \texttt{CVPR' 20} &28.2$\pm$0.3 & 53.0$\pm$0.1 & 52.7$\pm$0.1 & 78.7$\pm$0.2 & 68.6$\pm$0.1 & 87.7$\pm$0.1\\\cmidrule(){2-10}
     &  \multirow{5}{*}{Hard label} &Pseudo-Label \cite{lee2013pseudo}& \texttt{ICML' 13} &16.7$\pm$0.2 & 36.5$\pm$0.8 & 54.4$\pm$0.3 & 78.8$\pm$0.3 & 65.8$\pm$0.2 & 86.5$\pm$0.2\\
     & &Curriculum Pseudo-Label \cite{cascante2021curriculum} & \texttt{AAAI' 21} &20.5$\pm$0.5 & 41.7$\pm$0.5 & 53.4$\pm$0.8 & 78.3$\pm$0.5 & 69.1$\pm$0.3 & \myboxb{87.8$\pm$0.1}\\
     & &FixMatch \cite{sohn2020fixmatch}& \texttt{NIPS' 20} &\myboxb{28.1$\pm$0.1}  & \myboxb{51.8$\pm$0.6} & \textit{\myboxb{57.4$\pm$0.8}} & 78.5$\pm$0.5 & \myboxb{70.2$\pm$0.6} & 87.0$\pm$0.1\\
     & &FlexMatch \cite{zhang2021flexmatch}$^\ast$& \texttt{NIPS' 21} &27.3$\pm$0.5 & 49.7$\pm$0.8 & 53.4$\pm$0.2 & 77.9$\pm$0.3 & 67.6$\pm$0.5 & 87.0$\pm$0.2\\
     & &MoCo + FlexMatch$^\ast$ & \texttt{NIPS' 21} &\myboxc{35.0$\pm$1.2} & \myboxc{58.5$\pm$1.0} & {{53.4$\pm$0.4}} & 77.0$\pm$0.2 & {68.9$\pm$0.3} & 87.7$\pm$0.2\\\cmidrule(){2-10}
     &  \multirow{6}{*}{Soft label} & KD-Self-Training \cite{su2021realistic} & \texttt{CVPR' 21} &22.4$\pm$0.4 & 44.1$\pm$0.1 & 55.5$\pm$0.1 & {\myboxb{79.8$\pm$0.1}} & 67.7$\pm$0.2 & 87.5$\pm$0.2\\
     & &MoCo + KD-Self-Training  \cite{su2021realistic}& \texttt{CVPR' 21} &{31.9$\pm$0.1} & {56.8$\pm$0.1} & \myboxc{55.9$\pm$0.2} & \myboxc{80.3$\pm$0.1} & \myboxc{70.1$\pm$0.2} & \myboxc{88.1$\pm$0.1}\\   
      &   & SimMatch \cite{zheng2022simmatch}$^{\ast}$ & \texttt{CVPR' 22} &24.8$\pm$0.5 & 48.1$\pm$0.6 & 53.3$\pm$0.5 & 77.9$\pm$0.8 & 65.4$\pm$0.2 & 86.9$\pm$0.3\\
     & &MoCo + SimMatch$^{\ast}$ & \texttt{CVPR' 22} &{32.9$\pm$0.4} & {57.9$\pm$0.3} & 53.7$\pm$0.2 & 78.8$\pm$0.5 & 65.7$\pm$0.3 & 87.1$\pm$0.2\\   \cmidrule(){3-10}
     & &SoC & Ours   & \textit{\resb{31.3}{0.8}{11.4\%}}     & \textit{\resb{55.3}{0.7}{6.8\%}}     & {\resb{57.8}{0.5}{0.7\%}}      &  \textit{\resb{80.8}{0.5}{1.3\%}}      & \textit{\resb{71.3}{0.3}{1.6\%}}         & \textit{\resb{88.8}{0.2}{1.1\%}}                     \\
      & &MoCo + SoC  & Ours  & \textbf{\resb{39.3}{0.2}{12.3\%}}     & \textbf{\resb{62.4}{0.4}{6.7\%}}     & \textbf{\resb{58.0}{0.4}{3.8\%}}      & \textbf{\resb{81.7}{0.4}{1.7\%}}     & \textbf{\resb{70.8}{0.4}{1.0\%}}           & \textbf{\resb{88.9}{0.5}{0.9\%}}              \\ \midrule \midrule
      \multirow{13}{*}{Semi-Fungi} &  \multirow{2}{*}{---} & \textcolor{light-gray-2}{Supervised oracle} & --- & \textcolor{light-gray-2}{60.2$\pm$0.8 }&\textcolor{light-gray-2}{83.3$\pm$0.9 }&\textcolor{light-gray-2}{73.3$\pm$0.1 }&\textcolor{light-gray-2}{92.5$\pm$0.3 }&\textcolor{light-gray-2}{73.8$\pm$0.3 }&\textcolor{light-gray-2}{92.4$\pm$0.3}\\
     &  &MoCo \cite{he2020momentum} & \texttt{CVPR' 20} &33.6$\pm$0.2 & 59.4$\pm$0.3 & 55.2$\pm$0.2 & {82.9$\pm$0.2} & 52.5$\pm$0.4 & 79.5$\pm$0.2\\\cmidrule(){2-10}
     &  \multirow{5}{*}{Hard label} & Pseudo-Label \cite{lee2013pseudo} & \texttt{ICML' 13} &19.4$\pm$0.4 & 43.2$\pm$1.5 & 51.5$\pm$1.2 & 81.2$\pm$0.2 & 49.5$\pm$0.4 & 78.5$\pm$0.2\\
     & & Curriculum Pseudo-Label \cite{cascante2021curriculum} & \texttt{AAAI' 21} &31.4$\pm$0.6 & 55.0$\pm$0.6 & 53.7$\pm$0.2 & 80.2$\pm$0.1 & 53.3$\pm$0.5 & 80.0$\pm$0.5\\ 
     & & FixMatch \cite{sohn2020fixmatch} & \texttt{NIPS' 20} &32.2$\pm$1.0  & {57.0$\pm$1.2} & 56.3$\pm$0.5 & 80.4$\pm$0.5 & {58.7$\pm$0.7} & 81.7$\pm$0.2\\
     & &FlexMatch \cite{zhang2021flexmatch}$^\ast$ & \texttt{NIPS' 21} &{36.0$\pm$0.9} & {59.9$\pm$1.1} & {\myboxb{59.6$\pm$0.5}} & {\myboxb{82.4$\pm$0.5}} & \myboxb{60.1$\pm$0.6} & \myboxb{82.2$\pm$0.5}\\
     & &MoCo + FlexMatch$^\ast$ & \texttt{NIPS' 21} &\myboxc{44.2$\pm$0.6} & \myboxc{67.0$\pm$0.8} & {\myboxc{59.9$\pm$0.8}} & \myboxc{82.8$\pm$0.7} & \myboxc{61.4$\pm$0.6} & \myboxc{83.2$\pm$0.4}\\\cmidrule(){2-10}
     &  \multirow{6}{*}{Soft label} &  KD-Self-Training \cite{su2021realistic}  & \texttt{CVPR' 21} &{32.7$\pm$0.2} & 56.9$\pm$0.2 & {56.9$\pm$0.3} & {81.7$\pm$0.2} & 55.7$\pm$0.3 & {82.3$\pm$0.2}\\
     & & MoCo + KD-Self-Training \cite{su2021realistic} & \texttt{CVPR' 21} &{39.4$\pm$0.3} & {64.4$\pm$0.5} & {58.2$\pm$0.5 }& {84.4$\pm$0.2} & {55.2$\pm$0.5} & {82.9$\pm$0.2}\\
      &   & SimMatch \cite{zheng2022simmatch}$^{\ast}$ & \texttt{CVPR' 22} &\myboxb{36.5$\pm$0.9} & \myboxb{61.7$\pm$1.0} & 56.6$\pm$0.4 & 81.8$\pm$0.6  & 56.7$\pm$0.3 & 80.9$\pm$0.4\\
     & &MoCo + SimMatch$^{\ast}$ & \texttt{CVPR' 22} &{42.2$\pm$0.5} & {67.0$\pm$0.4} & 56.5$\pm$0.2 & {82.5$\pm$0.3} & 57.4$\pm$0.2 & 81.3$\pm$0.4\\   \cmidrule(){3-10}
     & &SoC & Ours   & \textit{\resb{39.4}{2.3}{7.9\%}}     & \textit{\resb{62.5}{1.1}{1.3\%}}     & \textit{\resb{61.4}{0.4}{3.0\%}}      &  \textit{\resb{83.9}{0.6}{1.8\%}}      & \textit{\resb{62.4}{0.2}{3.8\%}}          & \textit{\resb{85.1}{0.2}{3.5\%}}  \\     
      & &MoCo + SoC & Ours   & \textbf{\resb{47.2}{0.5}{6.8\%}}     & \textbf{\resb{71.3}{0.2}{6.4\%}}    & \textbf{\resb{61.9}{0.3}{3.3\%}}       &  \textbf{\resb{85.8}{0.2}{3.6\%}}    & \textbf{\resb{62.5}{0.4}{1.8\%}}          &   \textbf{\resb{84.7}{0.2}{1.8\%}}          \\ %
      \bottomrule \bottomrule
      \end{tabular}
  }
\end{table*}

\section{Experiment}
\label{sec:exp}
\subsection{Experimental Setup}
\subsubsection{Baselines}
We use various SSL baseline methods (Pseudo-labeling \cite{berg2013poof}, Curriculum Pseudo-Labeling \cite{cascante2021curriculum}, KD-Self-Training \cite{su2021realistic}, FixMatch \cite{sohn2020fixmatch}, FlexMatch \cite{zhang2021flexmatch} and SimMatch \cite{zheng2022simmatch}) and self-supervised learning baseline methods (MoCo \cite{he2020momentum} and MoCo + $X$) for comparisons. See Sec. \ref{app:id} of supplementary for more details. Specially, \textcolor{light-gray-2}{\textit{supervised oracle}} means the ground-truth labels of unlabeled data are included for training.
\subsubsection{Datasets}
\label{sec:dataset}
We evaluate SoC on two datasets that have been recently introduced into SS-FGCV: Semi-Aves and Semi-Fungi \cite{su2021semi,su2021realistic}. Both of them are consisting of 200 fine-grained classes and exhibit heavy class imbalance. Semi-Aves is divided into the training set and validation set with a total of 5,959 labeled images and 26,640
unlabeled images, and the test set with 8,000 images; Semi-Fungi is divided into the training set and validation set with a total of 4,141 labeled images and 13,166 unlabeled images, and the test set with 4,000 images. Additionally, Semi-Aves and Semi-Fungi also contain 800 and 1194 out-of-distribution (OOD) classes respectively, including 122,208 and 64,871 unlabeled images. SoC is mainly tested on Semi-Aves and Semi-Fungi with in-distribution data, while the two datasets with OOD data are also used for comprehensive evaluation.

\subsubsection{Implementation Details}
\label{sec:id}

For Semi-Aves and Semi-Fungi, all images are resized to 224 × 224. Following \citet{su2021realistic}, we adopt ResNet-50~\cite{he2016deep} as the backbone for all experiments including model training from scratch and pre-trained models on ImageNet \cite{deng2009imagenet} and iNaturalist 2018 (iNat) \cite{van2018inaturalist}, where iNat is a large-scale fine-grained dataset containing some overlapping classes with Semi-Aves (but there are no overlapping images). For consistency regularization,  following~\citet{sohn2020fixmatch},  crop-and-flip is used for weak augmentation and RandAugment~\cite{cubuk2020randaugment} is used for strong augmentation. The learning rate with cosine decay schedule is set to 0.01 for training from scratch and 0.001 for training from pre-trained models. For optimization, SGD is used with a momentum of 0.9 and a weight decay of 0.0005. For training from scratch, we train our models for 400$k$ iterations (200$k$ if MoCo is used), whereas our models are trained for 50$k$ iterations when training from pre-trained models.
For hyper-parameters, we set $N_{b}=5120, B=32, \mu =5,\lambda_{\mathtt{cos}}=1$ and $\alpha=5$. We report our results averaged on 3 runs with standard variances.

\newcommand{\mz}{4.2cm}
\newcommand{\wid}{0.24}
\begin{figure*}[t] 
    \vspace{-1em}
    \centering
    \subfloat[][SoC vs. FixMatch]{
    \includegraphics[width=\mz]{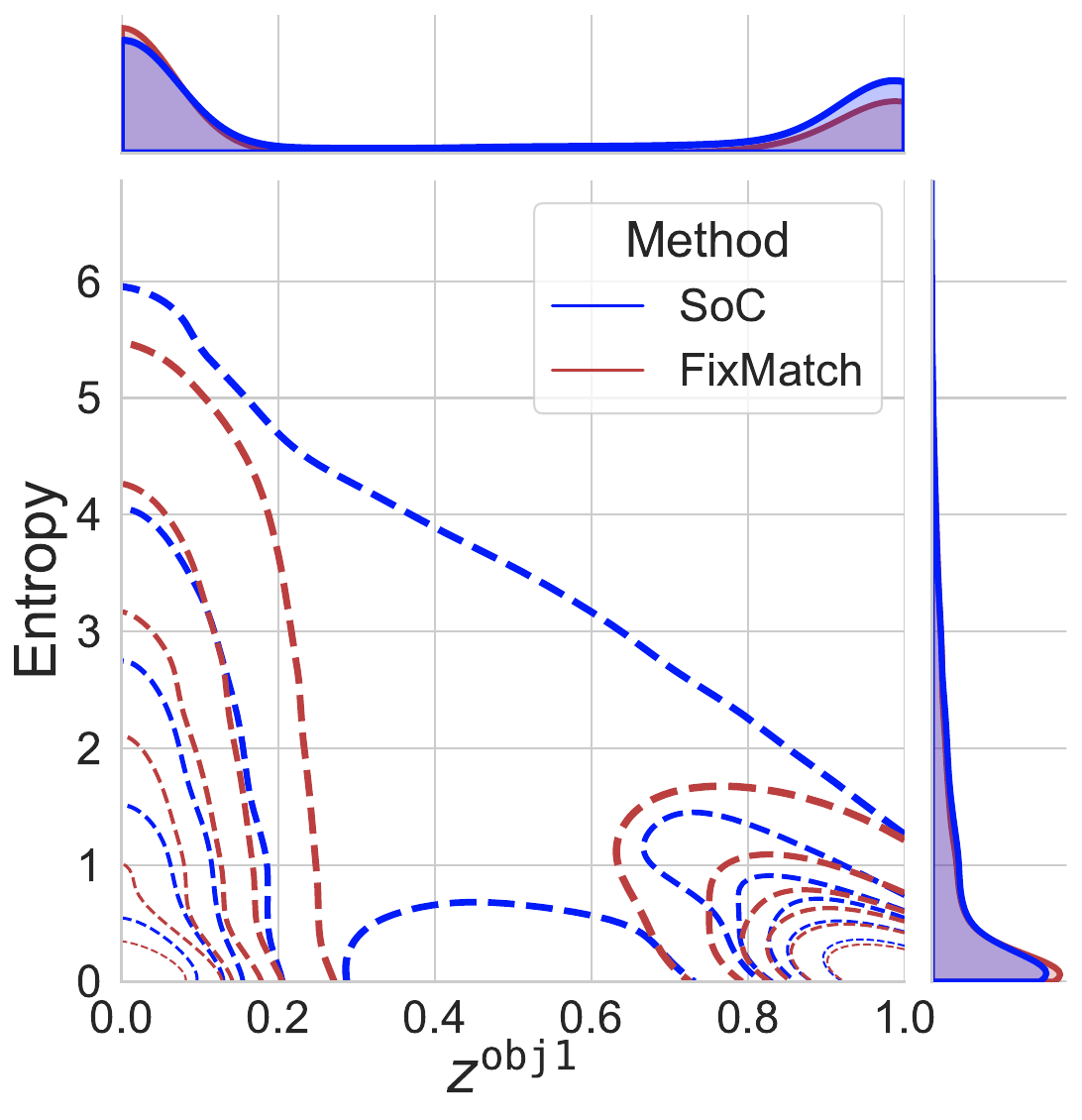}
    \label{fig:resa}
    }
    \subfloat[][SoC vs. Variant 1]{
    \includegraphics[width=\mz]{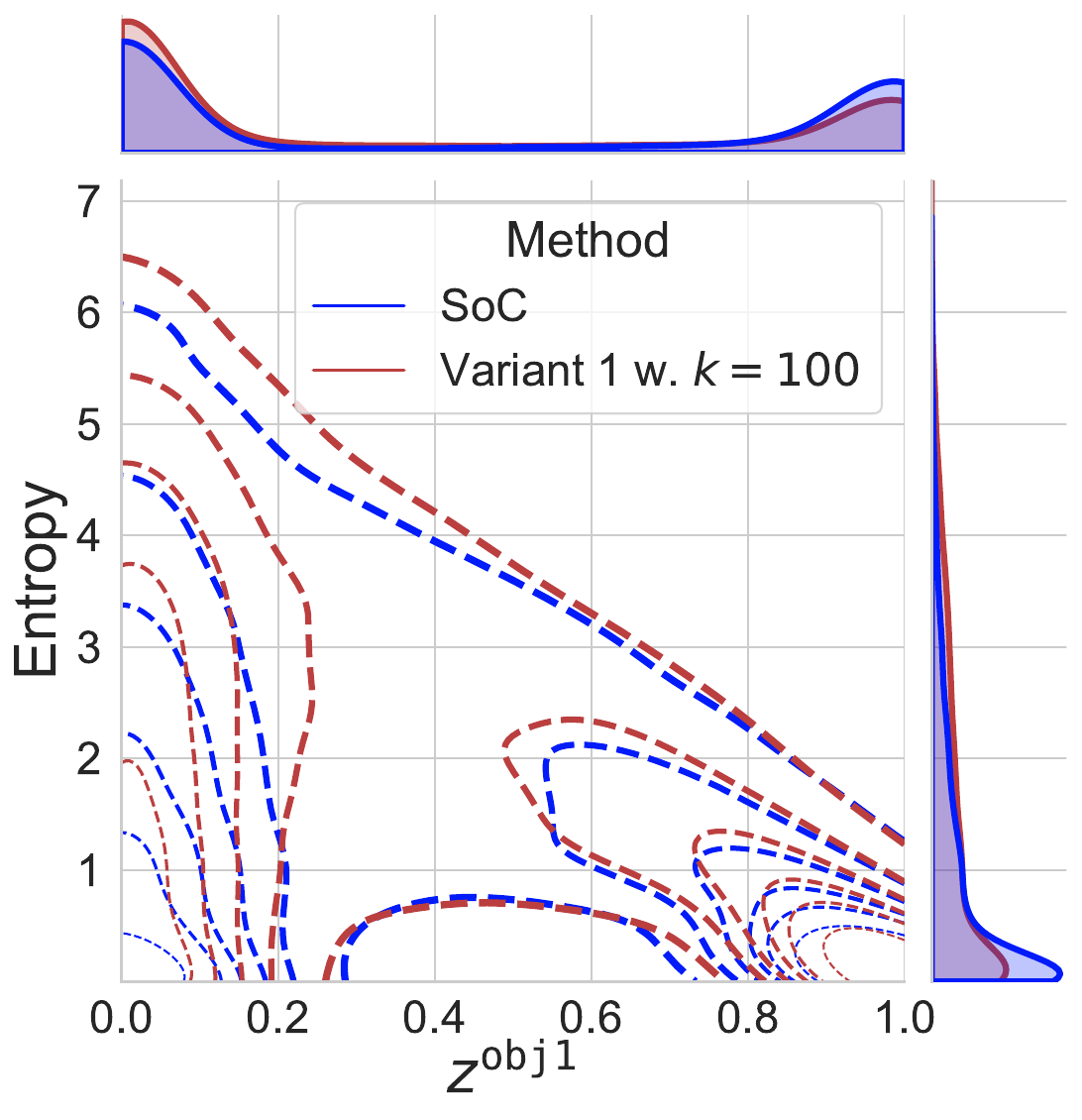}
    \label{fig:resb}
    }
    \subfloat[][SoC vs. Variant 2]{
    \includegraphics[width=\mz]{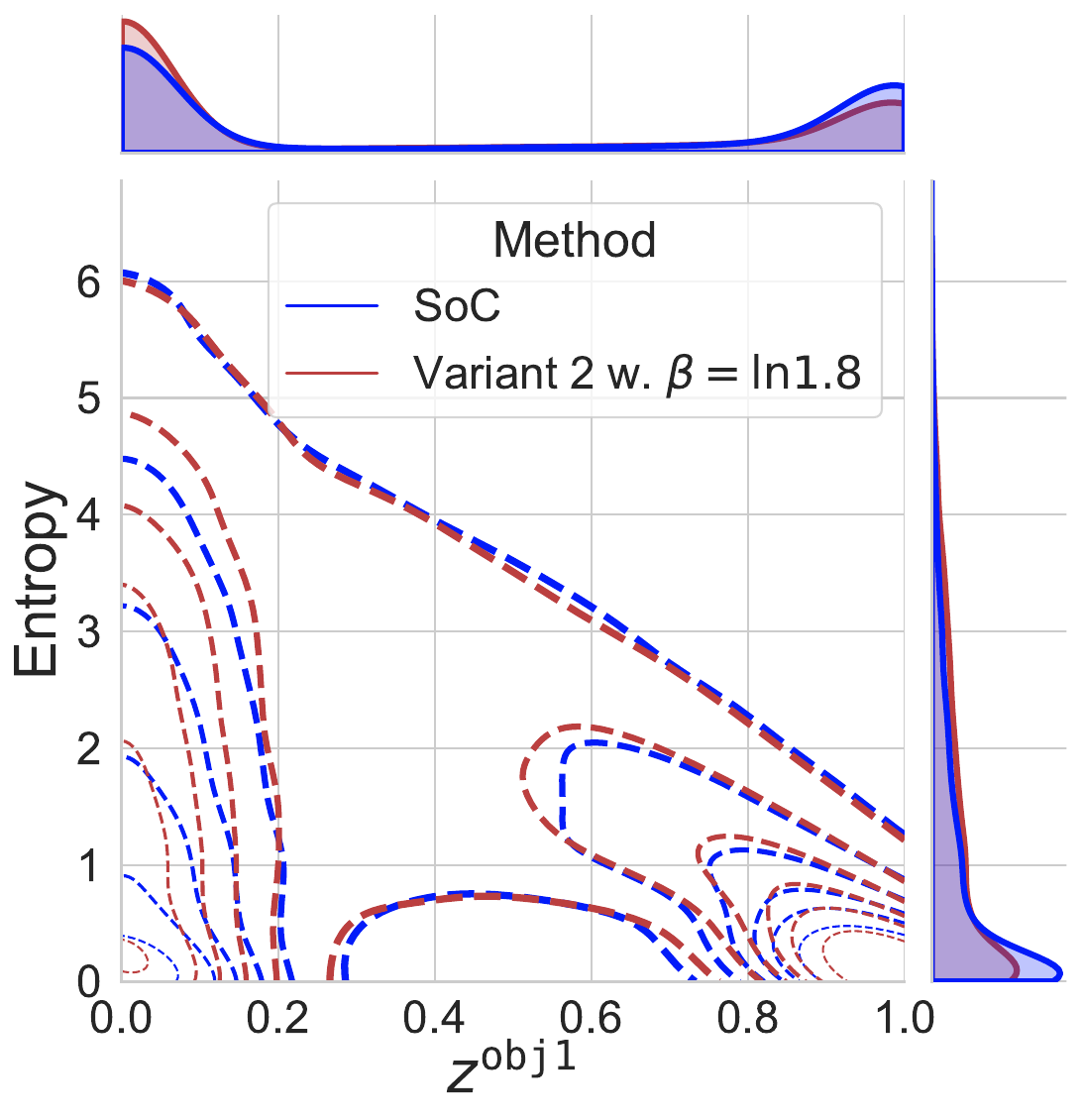}
    \label{fig:resc}
    }
    \subfloat[][SoC vs. Variant 3]{
    \includegraphics[width=\mz]{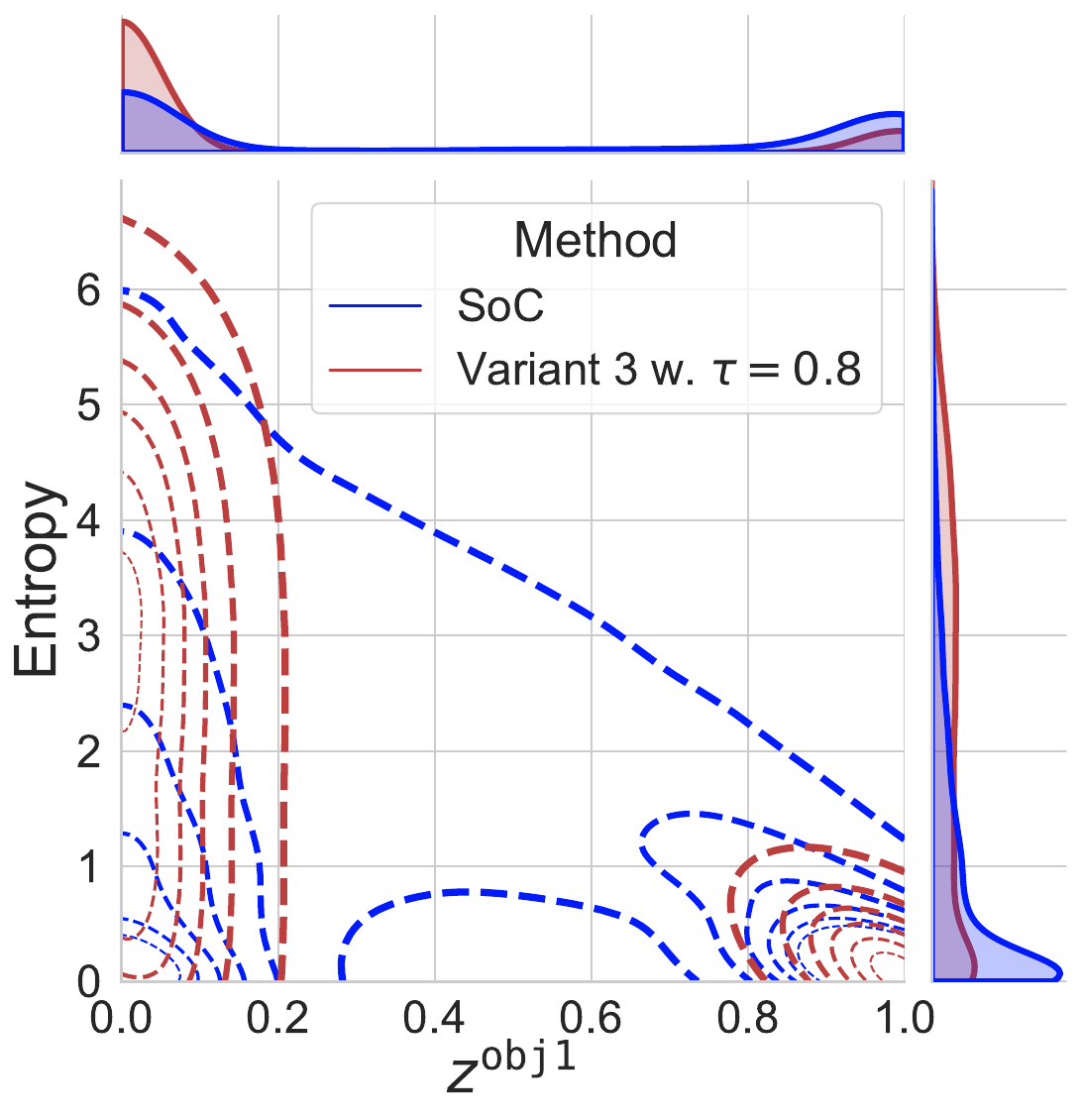}
    \label{fig:resd}
    }
    \caption{Kernel density estimation for the bivariate distribution of $z^{\mathtt{obj1}}$ and entropy. The x-axis represents $z^{\mathtt{obj1}}$, \ie, the optimization effect of Obj. \ref{eq:ob1}. The y-axis represents the entropy of predictions on the unlabeled data, \ie, the optimization effect of Obj. \ref{eq:ob2}. In general, the greater the density in the lower right corner of the region, the better the optimization effect. The experiments are conducted on Semi-Fungi training from scratch with the same experimental setting as Sec. \ref{sec:id}, In (a), we show the comparison between SoC and FixMatch.  In (b), (c) and (d),  we show the comparisons between SoC and the results of Variant 1, Variant 2 and Variant 3 respectively.}
    \label{fig:res}
  \end{figure*}
\subsection{Results and Analysis}
The comparisons of performance are summarized in Tab. \ref{table:1}. As shown in this table, SoC shows the most superior results on both Semi-Aves and Semi-Fungi. Without unsupervised learning (\ie, MoCo \cite{he2020momentum}), SoC outperforms all baseline SSL methods by a tangible margin.
Notably, from the perspective of hard pseudo-label based methods, FixMatch and FlexMatch underperform in SS-FGVC without pre-training. This confirms our claim that carefully selected soft labels are more suitable for SS-FGVC, because we can hardly guarantee the accuracy of pseudo-labels. Meanwhile, from the perspective of soft pseudo-label based methods, SimMatch's performance is also consistently weaker than SoC, which proves the effectiveness of our soft label selection scheme. The core enhancement of SoC is to optimize proposed \textit{coupled optimization goal}, which enables to generate more effective soft labels for SS-FGCV. As shown in Fig. \ref{fig:resa}, although SoC's predictions has a lower probability density than FixMatch on the low entropy region, it has higher probability density on high $z^{\mathtt{obj1}}$ regions, \ie, SoC ensures the attendance of ground-truth class in the pseudo-label at the slight expense of minimizing entropy, which leads to a substantial performance gain. With MoCo, SoC also consistently wins all baselines.

On the other hand, the performance of both baselines and SoC is boosted by pre-training, since the SSL model produces high-quality pseudo-labels at the outset. Even so, our method still shows the great
advantage of addressing FGVC. Additional results on Semi-Aves and Semi-Fungi with OOD data can be found in Sec. \ref{app:ood} of supplementary.

\subsection{Ablation Study}
\label{sec:ab}
To explore the effectiveness of each components in SoC, we mainly conduct experiments on three variants of SoC. More ablation studies on class transition tacking (CTT) based clustering can be found Sec. \ref{app:ab} of supplementary.

\noindent\textbf{Variant 1: Ablation of $k$ selection.} 
We retain the original CTT-based $k$-medoids clustering but use a fixed $k$ for it (\ie, ablating Confidence-Aware $k$ Selection). As show in Tab. \ref{tab:ab}, no matter what value  of $k$ is set, the default SoC achieves the best performance. In Fig. \ref{fig:resa}, we can observe that SoC achieves better results on our coupled optimization goal, \ie, compared to Variant 1, SoC's predictions are more clustered in regions with high $z^{\mathtt{obj1}}$ (corresponding to Obj. \ref{eq:ob1}) and low entropy (corresponding to Obj. \ref{eq:ob2}).

\noindent\textbf{Variant 2: Ablation of $f_{\mathtt{conf}}$.} First, we alter $\alpha$ for the default $f_{\mathtt{conf}}$ used in Eq. \eqref{eq:k} and the results are summarized in Tab. \ref{tab:ab}. Then, we use the exponential function to replace the liner function used for $f_{\mathtt{conf}}$ in default Confidence-Aware $k$ Selection, \ie, we rewrite Eq. \eqref{eq:k} as
\begin{equation}
    k_{i}=\lceil (\exp{(\beta\times\max(\mathbf{p}_{i}))-1+\frac{2}{K}}) \times K -\frac{1}{2} \rceil,
    \label{eq:k2}
\end{equation}
where $\beta\leq\ln(2-\frac{2}{K})$ (to keep $k_{i}\in\{2,\dots,K\}$). In  Tab. \ref{tab:ab} and  Fig. \ref{fig:resb}, we observe that the default SoC achieves better performance and optimization result, which proves that simple linear function is competent enough as $f_{\mathtt{conf}}$.

\noindent\textbf{Variant 3: Ablation of soft label selection.} 
We discard all components in SoC and use the most simple approach to utilize ordinary soft pseudo-label (\ie, $\mathbf{g}_{i}$ is set to an all-one vector) for learning in SS-FGVC, \ie, we replace the hard label used in FixMatch with soft label. As show in Tab. \ref{tab:ab}, with various confidence threshold $\tau$ in FixMatch, the performance of default SoC demonstrates the effectiveness of our proposed soft label selection. In addition, as shown in Fig. \ref{fig:resc}, we can observe that although the ordinary soft label contains all classes, Variant 3 still does not outperform SoC in optimizing $z^{\mathtt{obj1}}$. Meanwhile, the dynamic selection of $k$ has an extraordinary effect on entropy minimization, \ie, SoC's predictions are clustered in the low entropy region.

Additionally, the comparison between SoC and FixMatch in Fig. \ref{fig:resd} also demonstrated that a certain degree of sacrificing entropy minimization to make progress on optimizing Obj. \ref{eq:ob1} is more suitable for SS-FGVC. After all, it is necessary to output a ``certain'' prediction only after first ensuring that there is a correct answer in the pseudo-label.

\begin{table}[t]
   \centering
  \scriptsize
   \caption{Accuracy (\%) on Semi-Fungi ($K=200$) training from scratch. $k$, $\alpha$, $\beta$ and $\tau$ are altered for Variant 1, 2, 2 and 3 respectively. We mark the result of default SoC as \textbf{bold}.}
   \label{tab:ab}
\setlength{\tabcolsep}{2.7mm}{
  \begin{tabular}{@{}c|cccc@{}}
  \toprule\toprule
    $k$ & 25  & 50     & 100  &150       \\ \cmidrule(r){1-1}\cmidrule(lr){2-2} \cmidrule(lr){3-3} \cmidrule(lr){4-4} \cmidrule(lr){5-5} 
   Top-1 / Top-5&36.5 / 61.0 & 35.8 / 59.7 & 36.0 / 60.1 & 36.6 / 60.9   \\\cmidrule(){1-5}
    $\alpha$  & $K/(K-2)$  & 2     & 5  &10         \\ \cmidrule(r){1-1}\cmidrule(lr){2-2} \cmidrule(lr){3-3} \cmidrule(lr){4-4} \cmidrule(lr){5-5} 
   Top-1 / Top-5&36.6 / 60.8 & 37.8 / 61.2 & \textbf{39.4 / 62.5}  & 35.8 / 58.7  \\\cmidrule(){1-5}
   $\beta$  & $\ln{1.2}$  & $\ln{1.4}$     & $\ln{1.8}$  &$\ln{(2-\frac{2}{K})}$       \\ \cmidrule(r){1-1}\cmidrule(lr){2-2} \cmidrule(lr){3-3} \cmidrule(lr){4-4} \cmidrule(lr){5-5} 
   Top-1 / Top-5&34.9 / 58.4 & 36.7 / 60.2 & 37.2 / 60.3 & 36.9 / 60.2  \\\cmidrule(){1-5}
    $\tau$  & 0.2  & 0.4     & 0.8  & 0.95        \\\cmidrule(r){1-1}\cmidrule(lr){2-2} \cmidrule(lr){3-3} \cmidrule(lr){4-4} \cmidrule(lr){5-5} 
   Top-1 / Top-5&34.7 / 58.2 & 34.3 / 57.6 & 30.3 / 54.9 & 26.15 / 46.9   \\\bottomrule\bottomrule
   \end{tabular}}
\end{table}
\section{Conclusion}
We propose \textbf{So}ft Label Selection with Confidence-Aware \textbf{C}lustering based on Class Transition Tracking (SoC) to tackle the SS-FGVC scenario. SoC optimizes both \textit{Extension Objective} and \textit{Shrinkage Objective} in the tradeoff to improve the soft label selection for SS-FGVC. Comprehensive experiments show that SoC consistently achieves significant improvements over the current baseline methods in SS-FGVC. In the future, we believe our method can also be borrowed for more complex and realistic scenarios in SSL.

\section*{Acknowledgement}
This work is supported by the Science and Technology Innovation 2030 New Generation Artificial Intelligence Major Projects (SQ2023AAA010051), NSFC Program (62222604, 62206052, 62192783), Jiangsu Natural Science Foundation Project (BK20210224). In addition, we sincerely appreciate the discussion and helpful suggestions from Associate Professor Penghui Yao of Nanjing University.

\bibliography{aaai24}

\appendix
\onecolumn
\begin{center}
    \LARGE\bfseries Supplementary Material
\end{center}

\section{Algorithm}
\label{sec:alg}
\begin{algorithm*}[h]
    \small
    \textbf{Input}: similarity function $f_{\mathtt{sim}}$, class transition matrix $\mathbf{C}$, cluster number $k$
    \begin{algorithmic}[1]
      \caption{$k\text{-}\mathrm{medoids}_{f_{\mathtt{sim}}}$: Class Transition Tracking based $k$-medoids Clustering} %
      \label{a}
      \State $\{c^{(1)}_{1},\dots,c^{(1)}_{k}\}=\textrm{Initialize}(\mathcal{Y})$ \hfill\Comment{Randomly select centroids over the class space}
      \For{ $t=1$ \rm{\textbf{to}} $\mathrm{MaxIteration}$}
   \For{$i=1$ \rm{\textbf{to}} $k$} 
   
   \State $S^{(t)}_{i}=\{m\mid f_{\mathtt{sim}}(m,c^{(t)}_{i})\geq f_{\mathtt{sim}}(m,c^{(t)}_{j}),1\leq j\leq k\}$ \hfill\Comment{Assignment step} 
   \State \hfill \Comment{$f_{\mathtt{sim}}(m,n)=\frac{C_{mn}+C_{nm}}{2}$ is the similarity between class $m$ and $n$}
   \State \hfill \Comment{The larger the $f_{\mathtt{sim}}(m,n)$, the more similar the classes $m$ and $n$ are} 
   \For{$j=1$ \rm{\textbf{to}} $|S^{(t)}_{i}|$}
   \State $d^{(t)}_{j}=\sum^{|S^{(t)}_{i}|}_{l=1}f_{\mathtt{sim}}(m_{j},m_{l})$\hfill\Comment{$m_{j},m_{l}\in S^{(t)}_{i}$}
   \EndFor
   \State $c^{(t+1)}_{i}=\mathop{\arg\max}_{j\in\{1,\dots,k\}}(d^{(t)}_{j})$\hfill\Comment{Update medoids step}
   \EndFor
   \If{$\{c^{(t)}_{1},\dots,c^{(t)}_{k}\}=\{c^{(t+1)}_{1},\dots,c^{(t+1)}_{k}\}$}
   \State \textbf{return} $\mathcal{S}=\{S_{1},\dots,S_{k}\}$
   \EndIf
   \State \textbf{return} $\mathcal{S}=\{S_{1},\dots,S_{k}\}$
  \EndFor
\end{algorithmic}
\end{algorithm*}

\begin{algorithm*}[h]
    \small
    \textbf{Input}: class number: $K$, labeled data set: $\mathcal{D}^{\mathtt{lb}} =\{(\mathbf{x}_{i}^{\mathtt{lb}},\mathbf{y}_{i})\}^{N}_{i=1}$, unlabeled data set: $\mathcal{D}^{\mathtt{ulb}}=\{\mathbf{x}_{i}^{\mathtt{ulb}}\}^{M}_{i=1}$, model: $f_{\theta}$, weak and strong augmentation: $\mathtt{Aug}_{\mathtt{w}}$ and $\mathtt{Aug}_{\mathtt{s}}$, class prediction bank: $\{l_{i}\}^{M}_{i=1}$, class tracking matrices: $\{ \mathbf{C}^{(i)}\}^{N_{b}}_{i=1}$, CTT based $k$-medoids Clustering: $k\text{-}\mathrm{medoids}_{f_{\mathtt{sim}}}$,  mapping function from confidence to $k$: $f_{\mathtt{conf}}$   
    \begin{algorithmic}[1]
      \caption{SoC: \textbf{So}ft Label Selection with Confidence-Aware \textbf{C}lustering based on Class Transition Tracking} %
      \label{a2}
      \For{ $t=1$ \rm{\textbf{to}} $\mathrm{MaxIteration}$}
      \State Sample labeled data batch $\{(\lbi,\mathbf{y}_{i})\}^{B}_{i=1}\subset \mathcal{D}^{\mathtt{lb}}$
      \State Sample unlabeled data batch $\{\ulbi\}^{\mu B}_{i=1}\subset \mathcal{D}^{\mathtt{ulb}}$
      $\mathcal{L}_{\mathtt{sup}}=\frac{1}{B}\sum_{i = 1}^{B}H(\mathbf{y}_{i},f_{\mathtt{\theta}} (\lbi))$
      \hfill\Comment{Compute the supervised loss}
   \For{$i=1$ \rm{\textbf{to}} $\mu B$}
    \State $d=\mathrm{Index}(\ulbi)$ \hfill\Comment{Obtain the index of $\ulbi$}
    \State $\pwi=\mathrm{Softmax}(f_{\theta}(\mathrm{Aug}_{\mathtt{w}}(\ulbi)))$ \hfill\Comment{Compute soft pseudo-label for weakly-augmented $\ulbi$}
    
            \State $\hat{p}_{i}=\arg\max(\pwi)$ \hfill\Comment{Compute class prediction}
            \If{$l_{d}\neq \hat{p}_{i}$} 
            \State $C^{(n)}_{l_{d}\hat{p}_{i}}=C^{(n)}_{l_{d}\hat{p}_{i}}+1$ \hfill\Comment{Perform class transition tracking}
            $l_{d}=\hat{p}_{i}$
            \EndIf
\State $k_{i}=f_{\mathrm{conf}}(\max(\pwi))$ \hfill\Comment{Select $k$ for CTT based $k$-medoids Clustering} 
  \State $\left\{ S_{1},\dots,S_{k} \right\}=k\text{-}\mathrm{medoids}_{f_{\mathtt{sim}}}\left(\mathrm{Average}(\{ \mathbf{C}^{(i)}\}^{N_{b}}_{i=1}),k_{i}\right)$ \hfill\Comment{See Algorithm \ref{a}} 
  \State $\mathcal{C}_{i}=S_{s},s=\mathop{\arg\max}_{c\in\{1,\dots,k\}}\mathbbm{1}( \hat{p}_{i} \in S_{c})$ 
   \State $\textbf{g}_{i}=\left(\mathbbm{1}(1\in \mathcal{C}_{i}),\dots,\mathbbm{1}(K\in \mathcal{C}_{i})\right)$ \hfill\Comment{Compute label selection indicator $\mathbf{g}_{i}$} 
   \State $\Tilde{\mathbf{p}}_{i}=\mathrm{Normalize}(\mathbf{g}_{i}\circ\mathbf{p}_{i}) $ \hfill\Comment{Compute selected soft label} 
        
   \EndFor
   
   \State $\mathcal{L}_{\mathtt{cos}}=\frac{1}{\mu B}\sum_{i = 1}^{\mu B}H(\Tilde{\mathbf{p}}_{i}^{\mathtt{w}},f_{\theta}(\mathrm{Aug}_{\mathtt{s}}(\ulbi))) $\hfill\Comment{Compute the consistency loss}
         \State \textbf{return} $\mathcal{L} =\mathcal{L}_{\mathtt{sup}}+\lambda_{\mathtt{cos}} \mathcal{L}_{\mathtt{cos}} $\hfill\Comment{Optimize total loss}
      \EndFor
\end{algorithmic}
\end{algorithm*}

\section{Proof for Lemma \ref{the}}
\label{sec:proof}
In SoC, for $\ulbi$, we obtain the set of candidate classes $\mathcal{C}_{i}$, the prediction probabilities $\mathbf{p}_{i}$ and the selected soft label $\Tilde{\mathbf{p}}_{i}$. First, we resort the the vector sequence $\mathbf{p}_{i}=(p_{i,(1)},\dots,p_{i,(K)})$ to $(p_{i,(a_{1})},\dots,p_{i,(a_{|\mathcal{C}_{i}|})},p_{i,(b_{1})},\dots,p_{i,(b_{|\mathcal{Y}\setminus\mathcal{C}_{i}|})})$, where $K$ is the number of classes, $a_{j_{a}}\in\mathcal{C}_{i}$ and $b_{j_{b}}\in\mathcal{Y}\setminus\mathcal{C}_{i}$, \ie, we put the probabilities of unselected classes at the back end of the vector sequence. Meanwhile, the subsequences $(p_{i,(a_{1})},\dots,p_{i,(a_{|\mathcal{C}_{i}|})})$ and $(p_{i,(b_{1})},\dots,p_{i,(b_{|\mathcal{Y}\setminus\mathcal{C}_{i}|})})$ in $\mathbf{p}_{i}$ are sorted by value in descending order, \ie, $\max(p_{i,(a_{j_{a}})})=p_{i,(a_{1})}=p_{i,(1)}$ and $\max(p_{i,(b_{j_{b}})})=p_{i,(b_{1})}=p_{i,(\lci+1)}$. Then, denoting $d=\sum^{K}_{c=\lci+1}p_{i,(c)}$, we prove a equivalent form of Lemma \ref{the}:
\begin{align}
     -\sum^{K}_{c=1}p_{i,(c)}\log p_{i,(c)}& \geq -\sum^{\lci}_{c=1}\frac{p_{i,(c)}}{1-d}\log \frac{p_{i,(c)}}{1-d} \nonumber\\
      & =-\sum^{\lci}_{c=1}\frac{p_{i,(c)}}{1-d}\log p_{i,(c)} + \sum^{\lci}_{c=1}\frac{p_{i,(c)}}{1-d}\log(1-d) \nonumber\\
      & =-\sum^{\lci}_{c=1}\frac{p_{i,(c)}}{1-d}\log p_{i,(c)} + \log(1-d). \label{eq:new1}
\end{align}
Rewriting Eq. \eqref{eq:new1} we obtain its equivalent:
\begin{align}
    \sum^{\lci}_{c=1}\frac{p_{i,(c)}}{1-d}\log p_{i,(c)}-\sum^{\lci}_{c=1}p_{i,(c)}\log p_{i,(c)} &\geq \sum^{K}_{c=\lci+1}\ppi\log \ppi+\log (1-d) \nonumber\\
    (\frac{d}{1-d})\sum^{\lci}_{c=1}p_{i,(c)}\log p_{i,(c)} &\geq \sum^{K}_{c=\lci+1}\ppi\log \ppi + \log(1-d).\label{eq:newin1}
\end{align}
When $d=0$, Eq. \eqref{eq:newin1} obviously holds. Given $d=\sum^{K}_{c=\lci+1}p_{i,(c)}>0$, according to the property of $h(x)=x\log x$, when only one entry $p_{i,(z)}$ of  $(p_{i,(\lci+1)},\dots,p_{i,(K)})$ has a non-zero value, $\sum^{K}_{c=\lci+1}\ppi\log \ppi$ is maximized, \ie, the term on the right side of the inequality Eq. \eqref{eq:newin1} obtains the maximum. Given $\max(p_{i,(\lci+1)},\dots,p_{i,(K)})=p_{i,(\lci+1)}$, the non-zero entry $p_{i,(z)}$ is obviously $p_{i,(\lci+1)}$. In this case, we have $d=p_{i,(\lci+1)}$ and $\sum^{K}_{c=\lci+1}\ppi\log \ppi=p_{i,(\lci+1)} \log p_{i,(\lci+1)}$. Thus, we rewrite Eq. \eqref{eq:newin1} as

\begin{align}
    (\frac{p_{i,(\lci+1)}}{1-p_{i,(\lci+1)}})\sum^{\lci}_{c=1}p_{i,(c)}\log p_{i,(c)} &\geq p_{i,(\lci+1)} \log p_{i,(\lci+1)} + \log(1-p_{i,(\lci+1)})\nonumber\\
    (\frac{p_{i,(\lci+1)}}{1-p_{i,(\lci+1)}})(p_{i,(1)}\log p_{i,(1)}+\sum^{\lci}_{c=2}p_{i,(c)}\log p_{i,(c)}) &\geq p_{i,(\lci+1)} \log p_{i,(\lci+1)} + \log(1-p_{i,(\lci+1)}).\label{eq:newin2}
\end{align}
It's worth noting that we have $\arg\max(\mathbf{p}_{i})\in\mathcal{C}_{i}$ by Eq. \eqref{eq:s}, therefore we obtain $\max(\mathbf{p}_{i})\in\{p_{i,(1)},\dots,p_{i,(\lci)}\}$, \ie, $p_{i,(1)}\geq p_{i,(\lci+1)}$. In other words, there is no constraint on the value of $\sum^{\lci}_{c=2}p_{i,(c)}\log p_{i,(c)}$ when $p_{i,(1)}$ and $p_{i,(\lci+1)}$ are given. Thus, by Jensen Inequality we know that when $p_{i,(2)}=p_{i,(3)}=\dots=p_{i,(\lci)}=\frac{1-p_{i,(1)}-p_{i,(\lci+1)}}{\lci-1}$, $\sum^{\lci}_{c=2}p_{i,(c)}\log p_{i,(c)}$ obtains the minimum, \ie, the left side of the inequality Eq. \eqref{eq:newin2} obtains the minimum with the given $p_{i,(1)}$ and $p_{i,(\lci+1)}$ where $p_{i,(1)}\geq p_{i,(\lci+1)}$. In this case, we can rewrite Eq. \eqref{eq:newin2} as
\begin{align}
    \frac{p_{i,(\lci+1)}}{1-p_{i,(\lci+1)}}p_{i,(1)}\log p_{i,(1)}+  \frac{p_{i,(\lci+1)}(\lci-1)}{1-p_{i,(\lci+1)}}&\times\frac{1-p_{i,(1)}-p_{i,(\lci+1)}}{\lci-1}\log \frac{1-p_{i,(1)}-p_{i,(\lci+1)}}{\lci-1} \nonumber\\
    &\geq p_{i,(\lci+1)} \log p_{i,(\lci+1)} + \log(1-p_{i,(\lci+1)}). \label{eq:newin3}
\end{align}
By Eq. \eqref{eq:newin3}, we obtain
\begin{align}
    \frac{p_{i,(\lci+1)}}{1-p_{i,(\lci+1)}}p_{i,(1)}\log p_{i,(1)}+  \frac{p_{i,(\lci+1)}(1-p_{i,(1)}-p_{i,(\lci+1)})}{1-p_{i,(\lci+1)}}&[\log (1-p_{i,(1)}-p_{i,(\lci+1)})-\log(\lci-1)]\nonumber\\
    &\geq p_{i,(\lci+1)} \log p_{i,(\lci+1)} + \log(1-p_{i,(\lci+1)}) \nonumber\\
    \frac{p_{i,(\lci+1)}}{1-p_{i,(\lci+1)}}p_{i,(1)}\log p_{i,(1)}+  \frac{p_{i,(\lci+1)}(1-p_{i,(1)}-p_{i,(\lci+1)})}{1-p_{i,(\lci+1)}}&\log (1-p_{i,(1)}-p_{i,(\lci+1)})-\nonumber\\
    p_{i,(\lci+1)} \log p_{i,(\lci+1)} - \log(1-p_{i,(\lci+1)}) &\geq \frac{p_{i,(\lci+1)}(1-p_{i,(1)}-p_{i,(\lci+1)})}{1-p_{i,(\lci+1)}} \log(\lci-1)\nonumber\\
    \frac{p_{i,(1)}}{(1-p_{i,(1)}-p_{i,(\lci+1)})}\log p_{i,(1)}+ &\log (1-p_{i,(1)}-p_{i,(\lci+1)})-\nonumber\\
    \frac{(1-p_{i,(\lci+1)})(p_{i,(\lci+1)} \log p_{i,(\lci+1)} + \log(1-p_{i,(\lci+1)}))}{p_{i,(\lci+1)}(1-p_{i,(1)}-p_{i,(\lci+1)})} &\geq \log(\lci-1)
    .\label{eq:newin4}
\end{align}
Letting $x=p_{i,(1)}$ and $y=p_{i,(\lci+1)}$, we define the function 
\begin{equation}
    f(x,y)=\frac{x}{1-x-y}\log x+\log(1-x-y)-\frac{(1-y)(y\log y+\log(1-y))}{y(1-x-y)}.
\end{equation}
Then we can obtain
\begin{equation}
    \frac{\partial f(x,y)}{\partial x}= {{-\frac{(y-1)\left(y\log(x)-\log(1-y)-y\log(y)\right)}{y\left(1-x-y\right)^{2}}}}\geq 0.
\end{equation}
Given $x\geq y$ (due to $p_{i,(1)}\geq p_{i,(\lci+1)}$), $f(x,y)$ is minimized when $x=y$. Plugging $x=y$ into $f(x,y)$, we obtain
\begin{equation}
    f(y)=\frac{y}{1-2y}\log y+\log(1-2y)-\frac{(1-y)(y\log y+\log(1-y))}{y(1-2y)}
\end{equation}
and the first derivative of $f(y)$:
\begin{equation}
    f'(y)=\frac{(2y^{2}-4y+1)\log (1-y)}{(1-2y)^{2}y^{2}}.
\end{equation}
Solving $f'(y)=0$ we obtain $y=1-\frac{\sqrt{2}}{2}$. It is easy to verify that $f'(y)$ exists for all $y$ such that $0<y<\frac{1}{2}$ and the sign of $f'(y)$ changes from negative to positive. Thus, $f(y)$ has a minimum at $y=1-\frac{\sqrt{2}}{2}$, \ie, 
\begin{equation}
    \min f(y)=f(1-\frac{\sqrt{2}}{2})=\frac{1}{4}\left(\sqrt{2}\log(16)+\log(64)-4\log\left(1-\frac{\sqrt{2}}{2}\right)+4\log(\sqrt{2}-1)\right)\approx 2.36655.\label{eq:new5}
\end{equation}
Obviously, we have $\log{10}<f(1-\frac{\sqrt{2}}{2})<\log{11}$. By Eq. \eqref{eq:new5}, we know that Eq. \eqref{eq:newin4} holds when $\lci\leq 11$, \ie, $z^{\mathtt{obj2}}_{i}\leq 11$ (by $z^{\mathtt{obj2}}_{i}=\sum_{c=1}^{K} \gi$ and Eq. \eqref{eq:gi}). It is worth noting that the strict requirement for $z^{\mathtt{obj2}}_{i}$ is to ensure that Eq. \eqref{eq:newin4} holds in the worst case, where the class distribution of \textit{selected classes} excluding $\arg\max(p_{i})$ is ``uniform distribution'' (\ie, for any $a,b\in\mathcal{C}_{i}\land a,b\neq\arg\max(p_{i})$, $p_{i,(a)}=p_{i,(b)}$) and the class distribution of \textit{unselected classes} is ``one-hot distribution'' (\ie, $|\{p_{i,(z)}\mid p_{i,(z)}\neq 0 \land z\in\mathcal{Y}\setminus\mathcal{C}_{i}\}|=1$). In the vast majority of cases, any $z^{\mathtt{obj2}}_{i}$ will make Eq. \eqref{eq:newin4}. Meanwhile, given an apposite $\alpha$, thanks to Confidence-Aware $k$ Selection in Sec. \ref{sec:cas}, $z^{\mathtt{obj2}}_{i}\leq 11$ for the worst case also can be guaranteed after a few training iterations because the model becomes more confident on $\ulbi$ (implying a smaller $z^{\mathtt{obj2}}_{i}$).  

In fact, the mentioned extreme situation is almost impossible to occur in reality.  For specific, with training, the prediction distribution outputted by the model will tend to be non-uniform unless the model learns nothing. Thus, the requirement for $z^{\mathtt{obj2}}_{i}$ will be greatly relaxed. To verify this point, we show an intuitive example: even if the class distribution of selected classes is uniform distribution (\ie, for any $a,b\in\mathcal{C}_{i}$, $p_{i,(a)}=p_{i,(b)}$), Lemma \ref{the} holds no matter what value $z^{\mathtt{obj2}}_{i}$ takes.
\begin{example}[Uniform class distribution of selected classes]
Suppose $p_{i,(a)}=p_{i,(b)}$ for any $a,b\in\mathcal{C}_{i}$ (implying $p_{i,(1)}=p_{i,(2)}=\dots=p_{i,(\lci)}=\frac{\sum^{\lci}_{c=1}p_{i,(c)}}{\lci}$), Lemma \ref{the} holds.
\end{example}

Denoting $r=\sum^{\lci}_{c=1}p_{i,(c)}$, we prove a equivalent form of Lemma \ref{the}:
\begin{align}
     -\sum^{K}_{c=1}p_{i,(c)}\log p_{i,(c)}& \geq -\sum^{\lci}_{c=1}\frac{p_{i,(c)}}{r}\log \frac{p_{i,(c)}}{r} \nonumber\\
      & =-\sum^{\lci}_{c=1}\frac{p_{i,(c)}}{r}\log p_{i,(c)} + \sum^{\lci}_{c=1}\frac{p_{i,(c)}}{r}\log r \nonumber\\
      & =-\sum^{\lci}_{c=1}\frac{p_{i,(c)}}{r}\log p_{i,(c)} + \log r. \label{eq:ab1}
\end{align}
Rewriting Eq. \eqref{eq:ab1} we obtain its equivalent:
\begin{equation}
    (\frac{1}{r}-1)\sum^{\lci}_{c=1}p_{i,(c)}\log p_{i,(c)}  \geq\sum^{K}_{c=\lci+1}p_{i,(c)}\log p_{i,(c)} + \log r.\label{eq:in1}
\end{equation}
Noting that we have $\sum^{K}_{c=\lci+1}p_{i,(c)}=1-r$ by $\sum^{K}_{c=1}p_{i,(c)}=1$. Thus, we obtain  $p_{i,(z)}=1-r$ where $p_{i,(z)}$ is the unique non-zero entry of unselected class distribution  in the worst case mentioned above, \ie, $\sum^{K}_{c=\lci+1}p_{i,(c)}\log p_{i,(c)}=(1-r)\log (1-r)$. And since $p_{i,(1)}=p_{i,(2)}=\dots=p_{i,(\lci)}=\frac{\sum^{\lci}_{c=1}p_{i,(c)}}{\lci}$, to ensure that Eq. \eqref{eq:in1} holds, we obtain
\begin{align}
    (\frac{1}{r}-1)\lci\frac{r}{\lci}\log\frac{r}{\lci}& \geq(1-r)\log(1-r) + \log r \nonumber\\
    (\frac{1}{r}-1)r\log\frac{r}{\lci}-(1-r)\log(1-r) - \log r &\geq0 \nonumber\\
    -r\log r-(1-r)\log \lci-(1-r)\log(1-r)  &\geq0.\label{eq:in2}
\end{align}
Hence, we have
\begin{equation}
    \frac{\partial(-r\log r-(1-r)\log \lci-(1-r)\log(1-r) )}{\partial \lci}=\frac{r-1}{\lci}.
\end{equation}
Meanwhile, we have $\frac{r-1}{\lci}<0$ by $r<1$ and $\lci>0$. Thus, when $\lci$ is maximized, the left side of the inequality Eq. \eqref{eq:in2} is minimized. By Eq. \eqref{eq:s}, we know that $\arg\max(\mathbf{p})\in\mathcal{C}_{i}$, therefore we obtain $\max(\mathbf{p})\in\{p_{i,(1)},\dots,p_{i,(\lci)}\}$. When $p_{i,(1)}=p_{i,(2)}=\dots=p_{i,(\lci)}=\frac{r}{\lci}$, for any $p_{i,(j)}\in\{p_{i,(\lci+1)},\dots,p_{i,(K)}\}$, we have $p_{i,(j)}\leq\max(\mathbf{p})$ implying $p_{i,(z)}\leq\frac{r}{\lci}$ holds, \ie,  $1-r\leq\frac{r}{\lci}$. Thus, the maximum of $\lci$ is $\frac{r}{1-r}$.  Plugging $\lci=\frac{r}{1-r}$ into Eq. \eqref{eq:in2}, we obtain
\begin{equation}
    -r\log r-(1-r)\log (\frac{r}{1-r})-(1-r)\log(1-r)  \geq0.\label{eq:in3}
\end{equation}
We define $f(r)=-r\log r-(1-r)\log (\frac{r}{1-r})-(1-r)\log(1-r)$ and obtain the derivative of $f(r)$:
\begin{equation}
    f'(r)=-\frac{1}{r}-\log r+\log \frac{r}{1-r}+\log (1-r).\label{eq:der}
\end{equation}
By Eq. \eqref{eq:der}, we know that $f'(r)<0$ holds. Thus, $f(r)$ is minimized as $r$ converges to $1$. Solving this limit, we obtain
\begin{equation}
    \lim _{r \to 1}f(r)=0.
\end{equation}
Hence, Eq. \eqref{eq:in2} holds, implying Lemma \ref{the} holds.

\begin{figure*}[t] 
    \vspace{-1em}
    \centering
    \subfloat[][Semi-Aves]{
    \includegraphics[width=\mz]{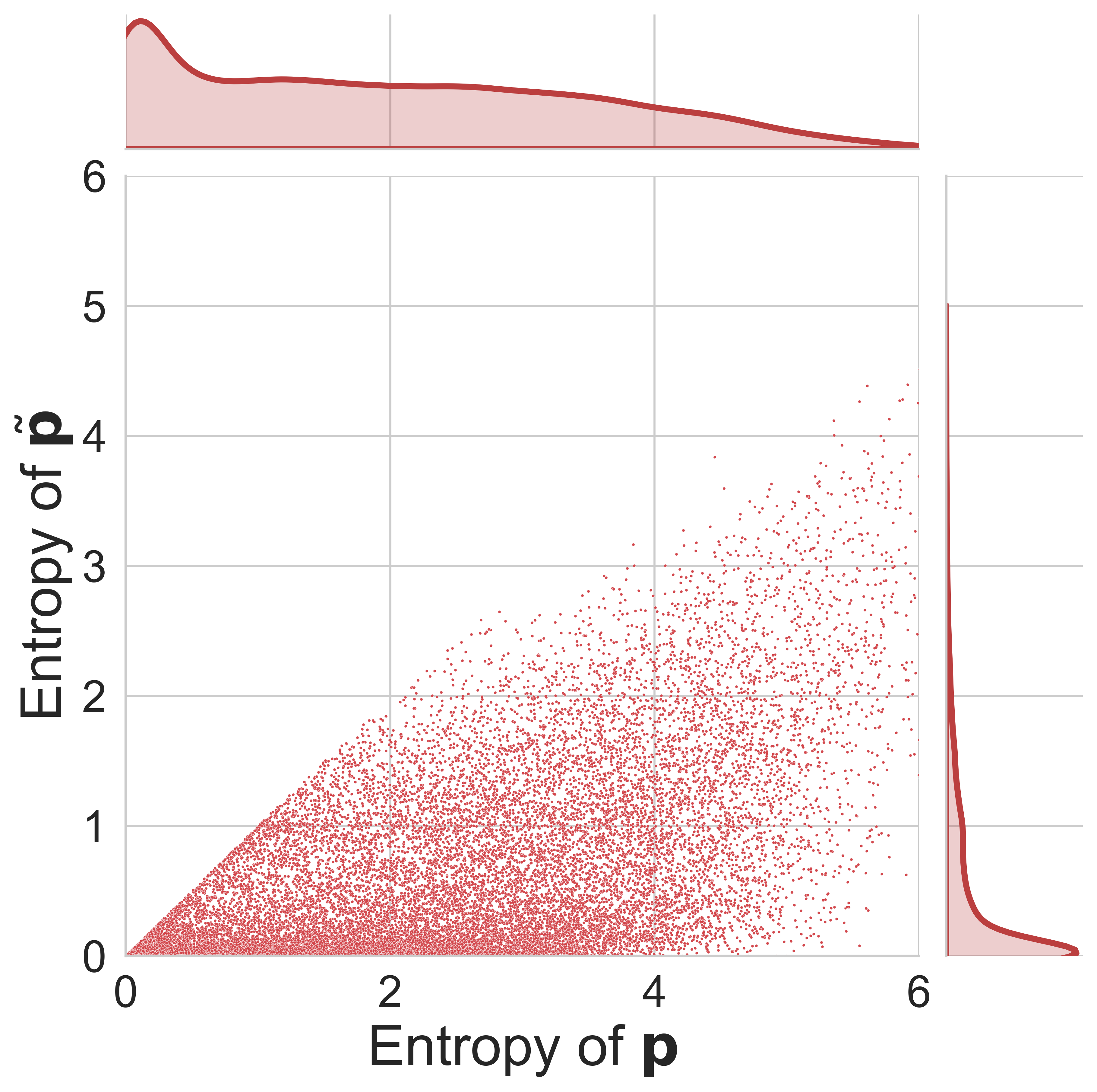}
    \label{fig:aresa}
    }
    \subfloat[][Semi-Aves]{
    \includegraphics[width=\mz]{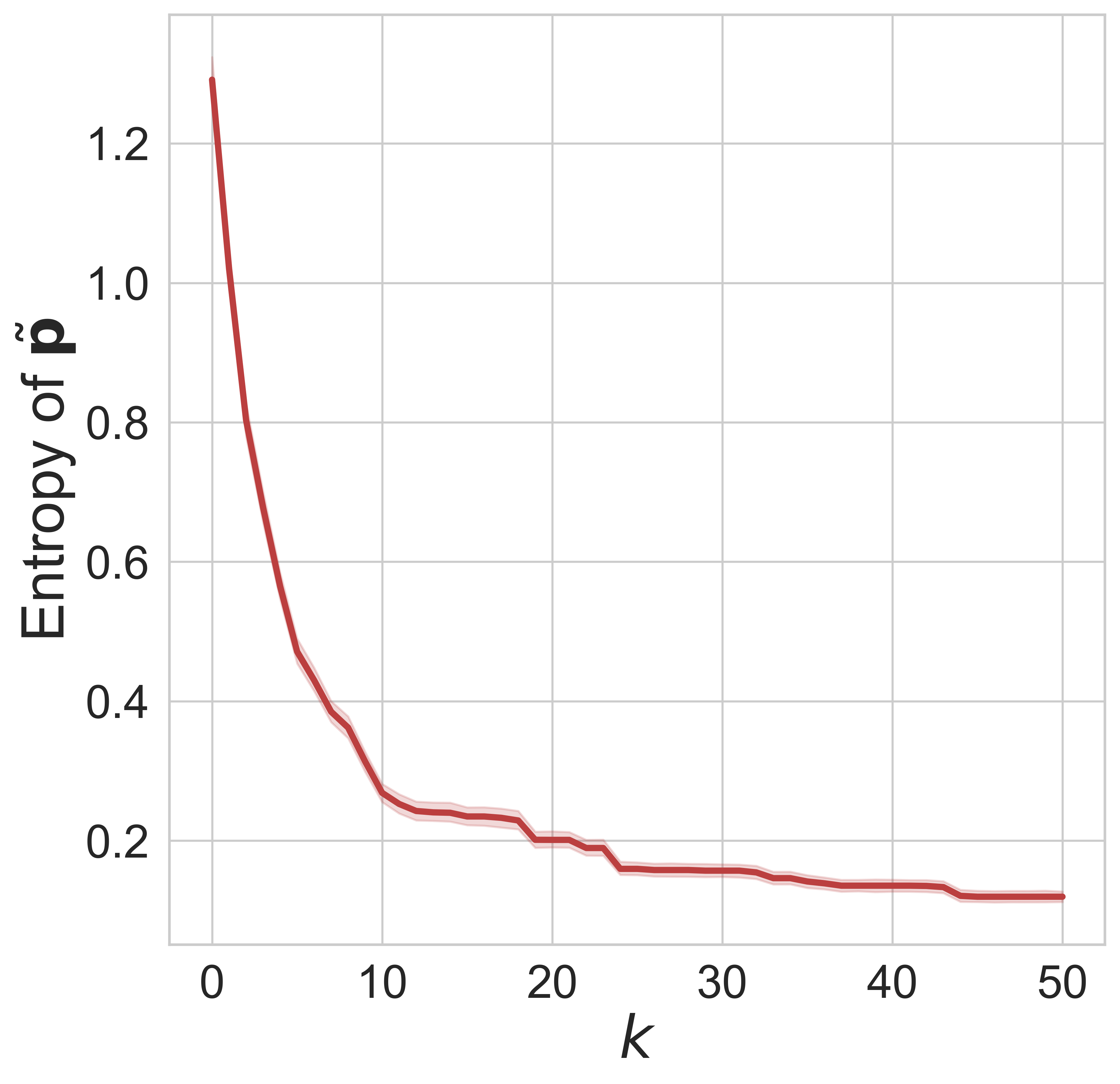}
    \label{fig:aresb}
    }
    \subfloat[][Semi-Fungi]{
    \includegraphics[width=\mz]{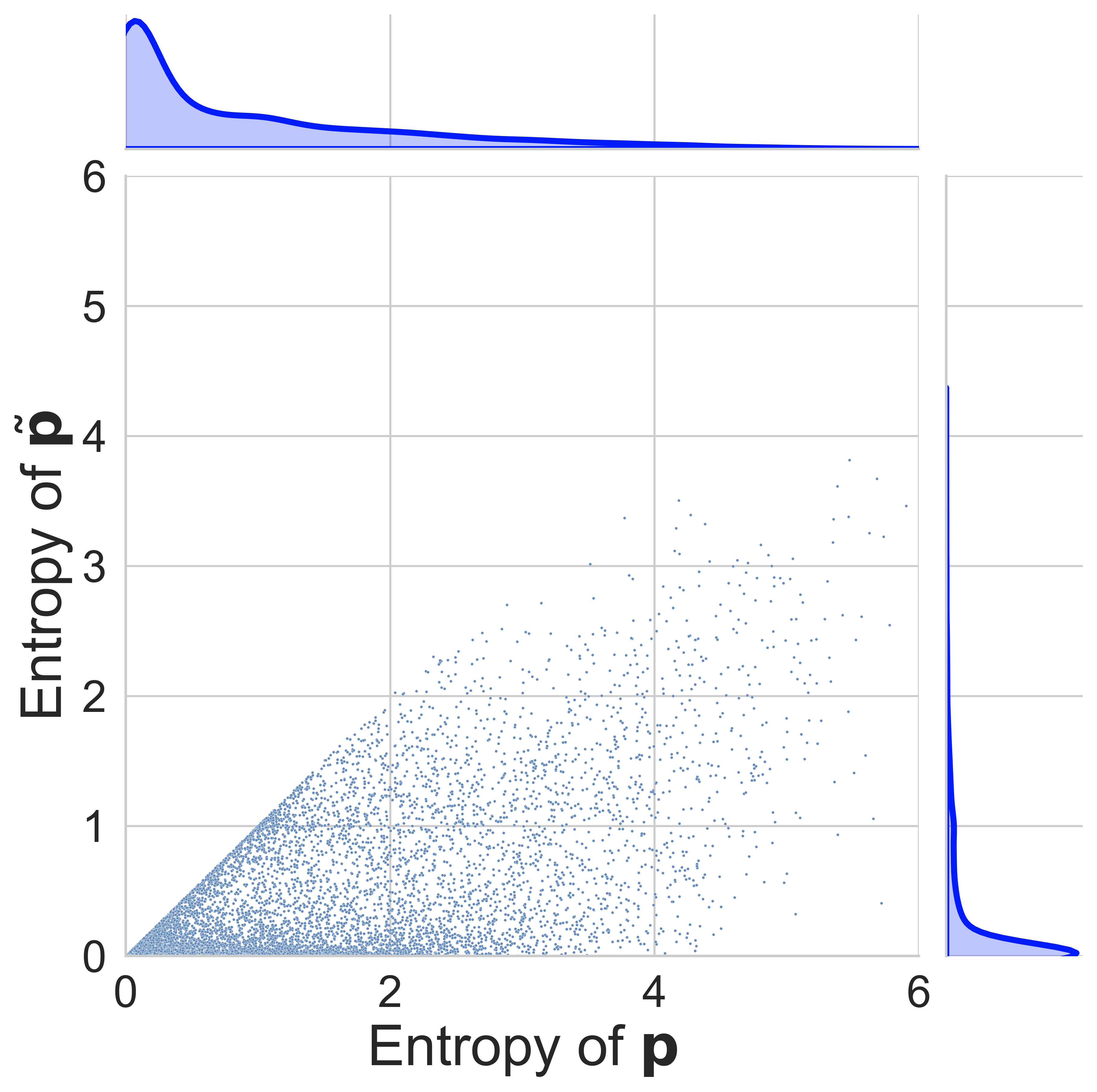}
    \label{fig:aresc}
    }
    \subfloat[][Semi-Fungi]{
    \includegraphics[width=\mz]{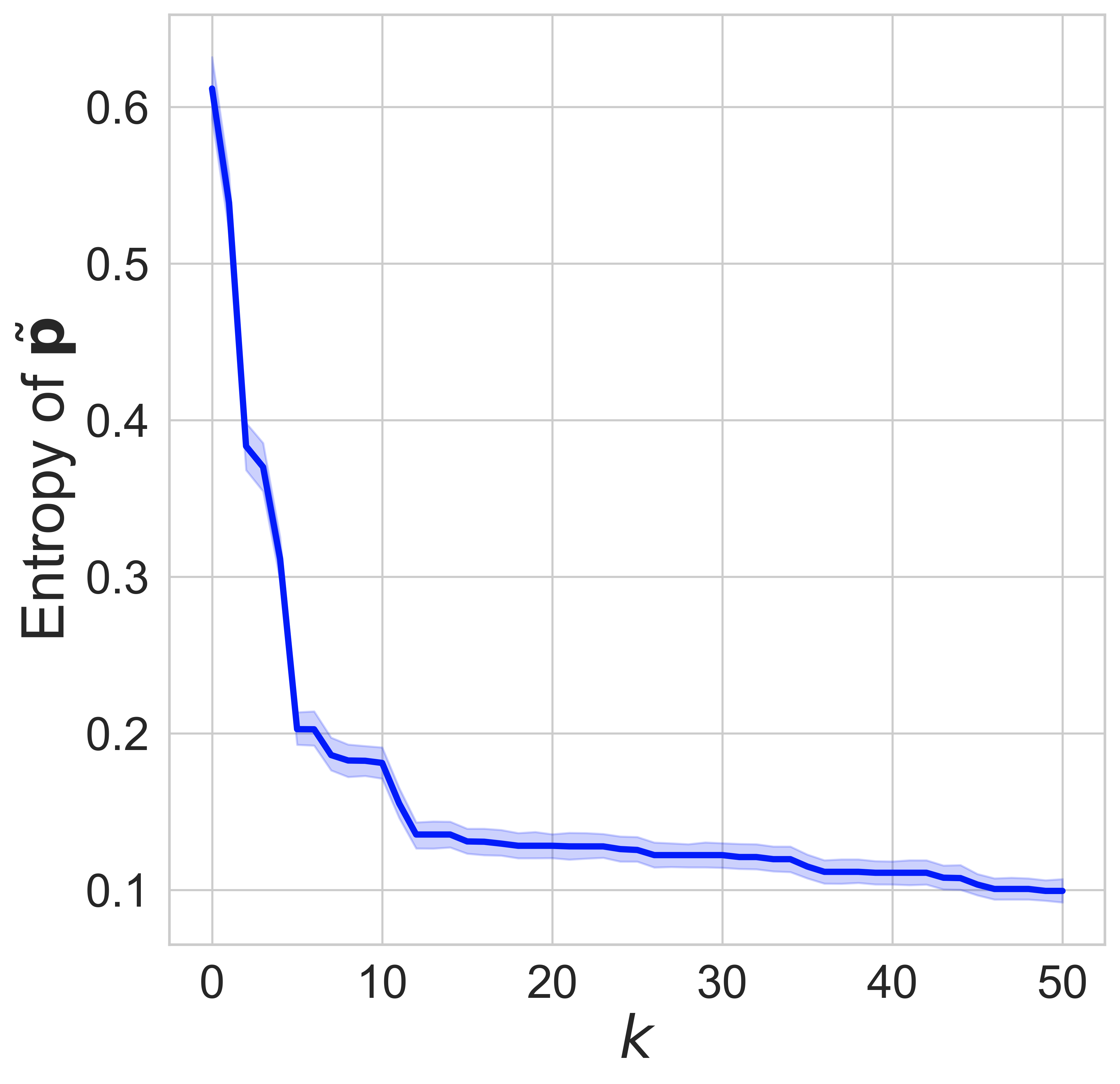}
    \label{fig:aresd}
    }
    \caption{(a) and (b): Visualization of tuple $(\mathcal{H}(\mathbf{p}_{i}),\mathcal{H}(\Tilde{\mathbf{p}}_{i}))$ with respective kernel density estimation ($\mathcal{H}(\cdot)$ refers to entropy). The scatter plots indicate the relative relationship between $\mathcal{H}(\mathbf{p}_{i})$ and $\mathcal{H}(\Tilde{\mathbf{p}}_{i})$, \ie, the distribution of data points below the diagonal line illustrates $\mathcal{H}(\Tilde{\mathbf{p}}_{i})<\mathcal{H}(\mathbf{p}_{i})$. (b) and (d): Visualization of the relationship between $\mathcal{H}(\Tilde{\mathbf{p}}_{i})$ and $k$. The results are averaged across all samples in the dataset and the 95\% confidence intervals with error bands are shown.}
    \label{fig:ares}
  \end{figure*}

On the other hand, even if the requirement for $z^{\mathtt{obj2}}_{i}$ is not met, Eq. \eqref{eq:newin4} still holds as long as the class distribution is not extremely close to the worst case above. In SoC, $|\mathcal{C}_{i}|$ is controlled by Confidence-Aware $k$ Selection to satisfy the requirement for $z^{\mathtt{obj2}}_{i}$ with a high probability while distribution characteristic of $\mathbf{p}_{i}$ is not close to the worst case discussed above. Thus, Eq. \eqref{eq:newin4} almost certainly holds in the training. We verify this through experiments in Fig. \ref{fig:ares}, while we observe that a larger $z^{\mathtt{obj2}}_{i}$ usually means a smaller entropy of $\Tilde{\mathbf{p}}_{i}$. In fact, Lemma \ref{the} holds for \textbf{all data points (\ie, 26.640 samples in Semi-Aves and 13,166 samples in Semi-Fungi)} in our experiments shown in Figs. \ref{fig:aresa} and \ref{fig:aresc}. In addition, as shown in Figs. \ref{fig:aresb} and \ref{fig:aresd}, the averaged entropy of $\Tilde{p}$ decreases rapidly with the increase of $k$, which generally reflects the strong role of Confidence-Aware $k$ Selection in optimizing Obj. \ref{eq:ob2}, yielding effective entropy minimization.
    
\section{Implementation Details}
\subsection{Comparison Methods}
\label{app:id}
Following \citet{su2021realistic},  baseline methods include various semi-supervised learning and self-supervised learning approaches for comprehensive comparisons.
\begin{itemize}[leftmargin=*]
    \item  \textbf{Supervised baseline.} The model is trained using the labeled data only. Specially, \textit{Supervised oracle} indicates the ground-truth of unlabeled are included for training.
       
    \item \textbf{Semi-supervised learning (SSL) method.} We adopt six representative SSL approaches for comparisons. (1) \textit{Pseudo-Labeling } \cite{lee2013pseudo}:  A approach utilizes the most confident predictions on the unlabeled data as hard labels to augment the training set. (2) \textit{Curriculum Pseudo-Labeling} \cite{cascante2021curriculum}: An improved pseudo-labeling method assigning pseudo-labels offline. 
(3) \textit{FixMatch} \cite{sohn2020fixmatch}: A prevailing approach unifying the consistency regularization and pseudo-labeling. It demonstrates the superior of using a confidence threshold to select more likely accurate pseudo-labels.  (4) \textit{FlexMatch} \cite{zhang2021flexmatch}: An improved FixMatch with curriculum pseudo labeling. (5) \textit{KD based Self-Training} \cite{su2021realistic}: A self-training method using knowledge distillation (KD) \cite{hinton2015distilling}. (6) \textit{SimMatch} \cite{zheng2022simmatch}: A soft pseudo-label framework simultaneously considering semantics similarity and instance similarity. In sum, (1)$\sim$(4) use hard pseudo-label and (5)$\sim$(6) use soft pseudo-label in the training.
\item \textbf{Self-supervised learning method.} (1) MoCo \cite{he2020momentum}: A momentum contrast based method for unsupervised representation learning. For SSL, MoCo is used for the pre-training on the unlabeled data followed by supervised fine-tuning on the labeled data. (2) MoCo + $X$: An unified approach using MoCo learning on the unlabeled data to initialize $X$'s model.
\end{itemize}

The implementation details of the methods except for FlexMatch \cite{zhang2021flexmatch} and SimMatch \cite{zheng2022simmatch} can be found in \citet{su2021realistic}. ResNet-50 is adopted as backbone for both FlexMatch and SimMatch, while $B$ (batch size) and $\mu$ (unlabeled data to labeled data ratio) is set to the identical value as what SoC uses for fair comparison. Specially, for
\textbf{FlexMatch:} with the hyper-parameter setting for ImageNet in the original paper, we use a learning rate of 0.003/0.03 to train models for 50k/400k iterations for training from expert models and from scratch; for
\textbf{SimMatch:} with the hyper-parameter setting for ImageNet in the original paper except for loss weight $\lambda_u$ and $\lambda_{in}$ are set to 1, we use a learning rate of 0.001/0.01 to train models for 800/1200 epochs for training from expert models and from scratch. We notice that SimMatch is more sensitive to hyper-parameter than FlexMatch. For example, although learning rate is set to 0.03 while $\lambda_u$ and $\lambda_{in}$ are set to 5 and 10 in the original paper of SimMatch \cite{zheng2022simmatch}, we find that they are too large,  resulting in severe performance degradation.
For MoCo + FlexMatch/SimMatch, we use the MoCo pre-trained models provided by \citet{su2021realistic} to initialize them. For details of MoCo + KD-Self-Training, please refer to \citet{su2021realistic}.
\subsection{Computational Cost}
\label{app:cc}
We provide more details for the computational overhead of clustering and class transition tracking. 
The clustering is done on the similarity matrix obtained from \textit{class transition matrix}, \ie, in the class space. Say, in the case of Semi-Fungi/Aves (200 classes), it is equivalent to clustering 200 sample points only, which incurs little overhead even if clustering is done per iteration. 
Also, CTT is only a loop at the time complexity of  $\mathcal{O}(n)$, \ie, it is performed once for each sample in the batch. On average, one training iteration takes 0.843s, while clustering takes 0.081s and CTT takes 0.047s. Compared with other models (\eg, FixMatch), the time cost does not increase tangibly. We use two GeForce RTX 3090s to train SoC, costing roughly 40G VRAM. Note that neither clustering nor CTT uses extra memory because we do not deploy them to GPU. 
In sum, SoC clearly wins naive soft label assignment (and other variants) in Tab. \ref{tab:ab} (\textbf{39.4 Top-1 accuracy}, and this is attained without incurring significant computational overhead.

\section{Additional Experimental Results}
\subsection{Empirical Evaluation on CTT-based Similiary Measure Properties}
\label{app:eec}
\begin{wrapfigure}{r}{7cm}
\vspace{-1em}
    \begin{minipage}{3cm}
    \centering\captionsetup[subfigure]{justification=centering}
    \subfloat[][Class transition matrix]{
    \includegraphics[width=3.2cm]{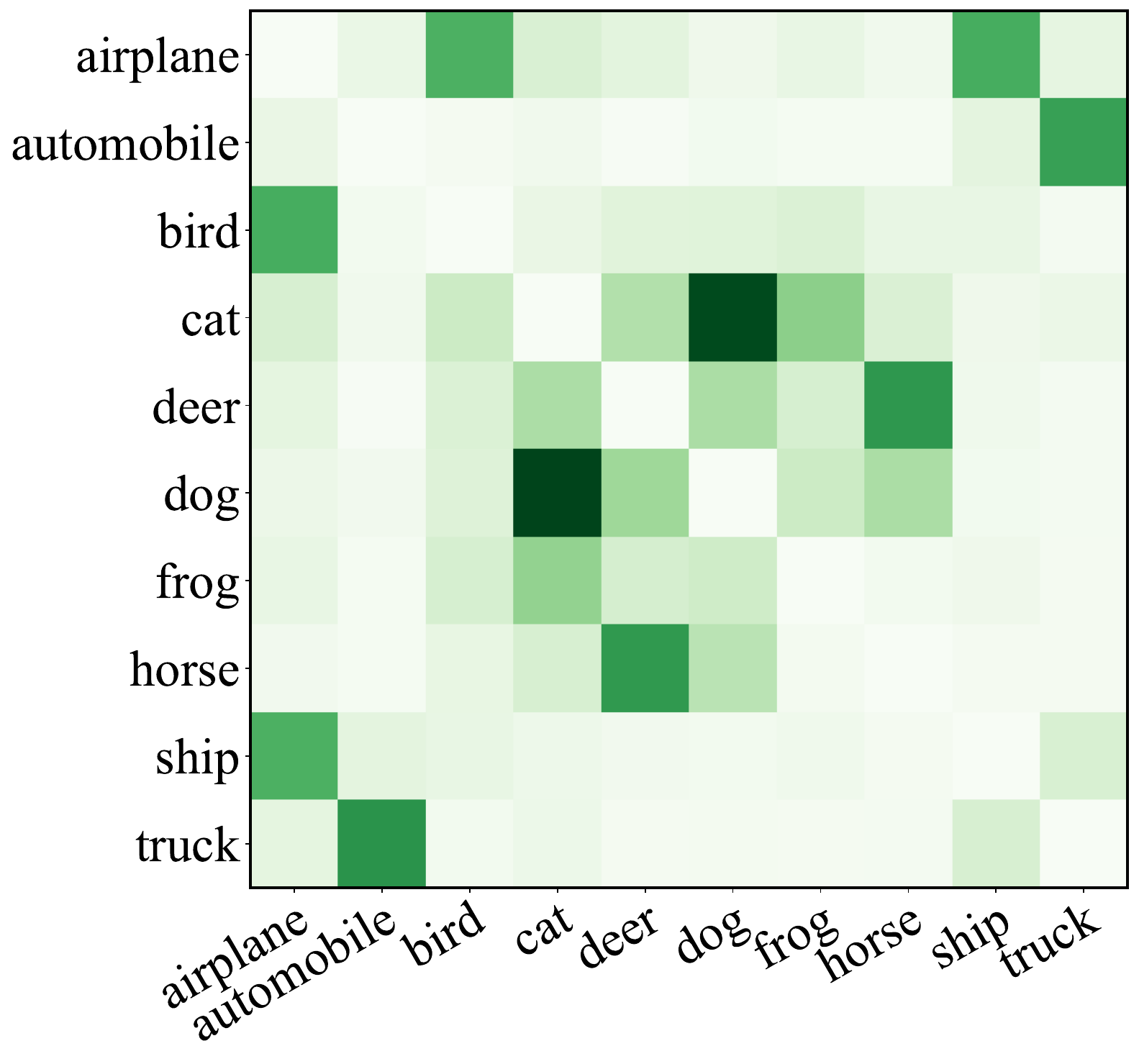}
    \label{app:ctt1}}
   \subfloat[][Confusion matrix]{
    \includegraphics[width=3.2cm]{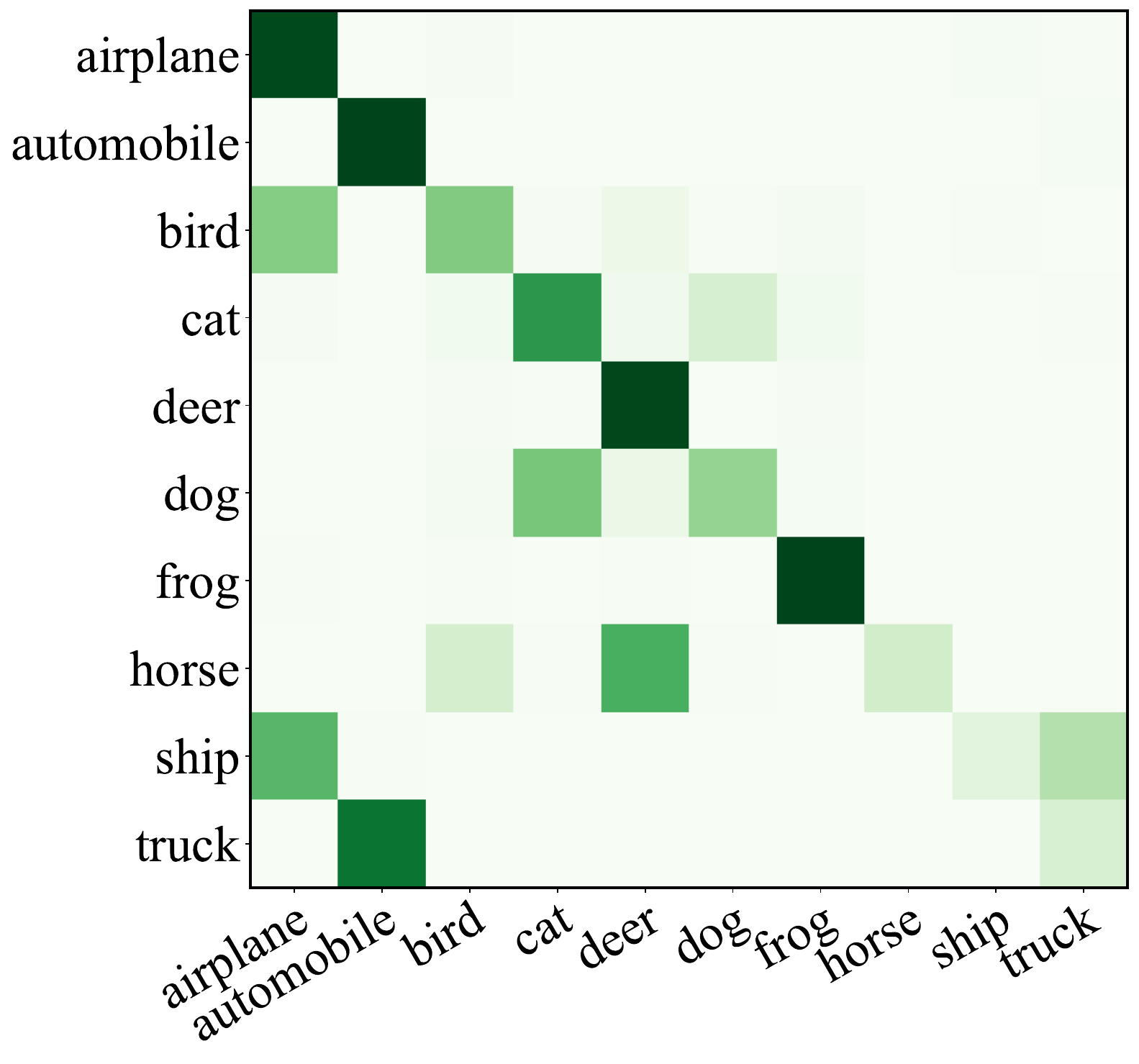}
    \label{app:ctt2}}

\end{minipage}
\caption{Empirical evaluation on CTT-based similarity measure properties (CIFAR-10). More frequent transitions between two classes mean that the two classes are more likely to be misclassified to each other, which means that the two classes are more similar.}\label{bare}
\vspace{-1em}
\end{wrapfigure}
In Sec. \ref{sec:cttc}, we claim that: the more frequent the transition between two classes, the greater the similarity between the two classes and the closer they are. For simplicity, we provide empirical evaluation on this measure of similarity provided by CTT  on {CIFAR-10}. As shown in Fig. \ref{bare}, \textit{class transitions} (\ref{app:ctt1}) and \textit{label mistakes} (\ref{app:ctt2}) occur more frequently between similar classes (mutually corroborated), \eg, ``dog'' and ``cat'', ``truck'' and ``automobile''. We can observe that the more similar the two classes are, the easier it is for the model to classify them as each other, and at the same time, the model swing more between them (resulting in more class transitions).

\subsection{More Results in OOD Setting}
\label{app:ood}
In this section, we provide more results on Semi-Fungi with OOD unlabeled samples to further demonstrate the superior of proposed SoC. As shown Tabs. \ref{table:2} and \ref{table:22}, SoC consistently outperforms baseline methods in OOD setting with training from pre-trained model, showing the potential of SoC for addressing more complex SS-FGVC scenarios.
\begin{wraptable}{r}{7cm}
  \scriptsize
    \centering
      \begin{tabular}{@{}l|ccc@{}}     
      \toprule
      Method & LD & PL & CTT \\  
      \midrule         
      Top-1 / Top-5   & 34.5 / 56.8   & 35.5 / 57.6      & 39.4 / 62.5  \\     
      \bottomrule
      \end{tabular}
\caption{
Accuracy (\%) on Semi-Fungi with training from scratch. \textit{LD} and \textit{PL} represent clustering by calculating the centroid of each class using the features of labeled data and pseudo-label, respectively.
  }
   \label{table:clu}
\end{wraptable}
\subsection{Ablation studies on CTT-based Clustering}
\label{app:ab}
To explain the design of CTT-based clustering, we illustrate the problem by using instance features commonly used in deep clustering. The mentioned instance similarity is not directly applicable for SoC, as our design is to cluster in class space rather than instance space. Therefore, using class transition matrix to directly depict class-level similarity is intuitive. Furthermore, we design two approaches based on instance feature similarity in Tab. \ref{table:clu} for comparison: we aggregate the centroids of the labeled data (LD) / pseudo-labels (PL) by class, and then cluster them in the class space. SoC wins the two alternatives because CTT-based clustering incorporates the historical and high-level information extracted from the learning of unlabeled data. 
\begin{table}[h]
\vspace{0.5em}
  \scriptsize
      \centering
      \caption{Results of accuracy (\%) on Semi-Aves and Semi-Fungi with out-of-distribution (OOD) classes in the unlabeled data. 
      The models are trained from scratch with the same settings in  Tab. \ref{table:1}. The results of baselines methods are  derived from \citet{su2021realistic}
  }
      ~\begin{tabular}{@{}l|cc|cc@{}}     
      \toprule\toprule
      \multirow{2}{*}{Method} & \multicolumn{2}{c|}{Semi-Aves} & \multicolumn{2}{c}{Semi-Fungi}  \\   
          & Top-1 & Top-5   &Top-1 & Top-5   \\ \midrule
    Supervised baseline &20.6$\pm$0.4 & 41.7$\pm$0.7 & 31.0$\pm$0.4 & 54.7$\pm$0.8\\
    MoCo \cite{he2020momentum} &38.9$\pm$0.4 & 65.4$\pm$0.3 & 44.6$\pm$0.4 & 72.6$\pm$0.5\\ \midrule
      Pseudo-Label \cite{lee2013pseudo} &12.2$\pm$0.8 & 31.9$\pm$1.6 & 15.2$\pm$1.0 & 40.6$\pm$1.2 \\
      Curriculum Pseudo-Label \cite{cascante2021curriculum} &20.2$\pm$0.5 & 41.0$\pm$0.9 & 30.8$\pm$0.1 & 54.4$\pm$0.3\\
     FixMatch \cite{sohn2020fixmatch} &19.2$\pm$0.2 & 42.6$\pm$0.6 & 25.2$\pm$0.3 & 50.2$\pm$0.8\\
     FlexMatch \cite{zhang2021flexmatch} &\myboxb{24.1$\pm$0.7} & \myboxb{47.5$\pm$1.0} & \myboxb{34.5$\pm$0.5} & \myboxb{57.1$\pm$1.1}\\
     MoCo + FlexMatch &{{39.4$\pm$0.4}} & {{63.8$\pm$0.6} }& {45.1$\pm$0.3} & {73.2$\pm$0.4} \\  \midrule
     Self-Training \cite{su2021realistic} &{22.0$\pm$0.5} & {43.3$\pm$0.2} & {32.5$\pm$0.5} & {56.3$\pm$0.3}\\
     MoCo + KD-Self-Training \cite{su2021realistic}&{\myboxc{41.2$\pm$0.2}} & {\myboxc{65.9$\pm$0.3} }& \myboxc{48.6$\pm$0.3} & \myboxc{74.7$\pm$0.2} \\   
     SimMatch \cite{zheng2022simmatch} &{21.1$\pm$0.8} & {43.6$\pm$0.9} & 20.6$\pm$0.5 & 43.4$\pm$0.7\\
     MoCo + SimMatch &{35.6$\pm$0.4} & {62.3$\pm$0.5}& 40.3$\pm$0.2& 69.0$\pm$0.5 \\  \midrule 
     SoC    & \textit{\tol{26.0}{1.0}{7.9\%}}         & \textit{\tol{49.1}{0.8}{3.4\%}}      &  \textit{\tol{35.4}{0.9}{2.3\%}}      &  \textit{\tol{58.3}{1.7}{2.1\%}}          \\
      MoCo + SoC    & \textbf{\tol{41.9}{0.6}{1.7\%}}     & \textbf{\tol{66.8}{0.2}{1.4\%}}     & \textbf{\tol{50.5}{0.4}{3.9\%}}      & \textbf{\tol{75.2}{0.3}{0.7\%}}         \\% \hline 
      \bottomrule\bottomrule
      \end{tabular}
  \label{table:2}
\end{table}

\begin{table}[h]
  \scriptsize
      \centering 
      \caption{Results of accuracy (\%) on Semi-Fungi with out-of-distribution (OOD) classes in the unlabeled data. 
      The models are trained from pre-trained models with the same settings in Tab. \ref{table:1}. The results of baselines methods are  derived from \citet{su2021realistic}.
  }
      ~\begin{tabular}{@{}l|cc|cc@{}}     
      \toprule\toprule
      \multirow{2}{*}{Method} & \multicolumn{2}{c|}{from ImageNet} & \multicolumn{2}{c}{from iNat}  \\   
          & Top-1 & Top-5   &Top-1 & Top-5   \\ \midrule
    Supervised baseline &53.8$\pm$0.4 & 80.0$\pm$0.4 & 52.4$\pm$0.6 & 79.5$\pm$0.5\\
    MoCo \cite{he2020momentum} &52.9$\pm$0.3 & 82.1$\pm$0.1 & 51.0$\pm$0.2 & 78.5$\pm$0.3\\ \midrule
      Pseudo-Label \cite{lee2013pseudo} &52.4$\pm$0.2 &80.4$\pm$0.5 &49.9$\pm$0.2& 78.5$\pm$0.3 \\
      Curriculum Pseudo-Label \cite{cascante2021curriculum} &54.2$\pm$0.2 &79.9$\pm$0.2 &53.6$\pm$0.3 &79.9$\pm$0.2\\
     FixMatch \cite{sohn2020fixmatch} &51.2$\pm$0.6 &77.6$\pm$0.3 &53.1$\pm$0.8 &79.9$\pm$0.1\\
     Self-Training \cite{su2021realistic} &\myboxb{55.7$\pm$0.3}& \myboxb{81.0$\pm$0.2}& \myboxb{55.2$\pm$0.2}& \myboxb{82.0$\pm$0.3}\\
     MoCo + KD-Self-Training \cite{su2021realistic} &{\myboxc{55.9$\pm$0.1}} & {\myboxc{82.9$\pm$0.2} }& \myboxc{54.0$\pm$0.2} & \myboxc{81.3$\pm$0.3} \\   \midrule
     SoC    & \textit{\tol{58.1}{0.5}{4.3\%}}         & \textit{\tol{82.3}{0.2}{1.6\%}}      &  \textit{\tol{58.9}{0.3}{6.7\%}}      &  \textit{\tol{82.6}{0.2}{0.7\%}}          \\
      MoCo + SoC    & \textbf{\tol{59.4}{0.2}{6.3\%}}     & \textbf{\tol{83.6}{0.1}{0.8\%}}     & \textbf{\tol{58.6}{0.2}{8.5\%}}      & \textbf{\tol{82.6}{0.1}{1.6\%}}         \\% \hline 
      \bottomrule\bottomrule
      \end{tabular}
  \label{table:22}
\end{table}

\end{document}